\documentclass{article}

\usepackage[preprint,nonatbib]{neurips_2020}

\usepackage[utf8]{inputenc} 
\usepackage[T1]{fontenc}    
\usepackage{hyperref}       
\usepackage{url}            
\usepackage{booktabs}       
\usepackage{amsfonts}       
\usepackage{nicefrac}       
\usepackage{microtype}      
\usepackage{xcolor}
\usepackage{footnote}
\usepackage{subcaption}
\usepackage{mwe}
\usepackage{graphicx}
\usepackage{pgf,pgfplots,pgfplotstable}
\usepackage{arydshln}
\usepackage{chngcntr}
\counterwithin{table}{section}
\graphicspath{ {./images/} }

\definecolor{butter}{rgb}{0.988,0.914,0.310}
\definecolor{chocolate}{rgb}{0.914,0.725,0.431}
\definecolor{chameleon}{rgb}{0.541,0.886,0.204}
\definecolor{skyblue}{rgb}{0.447,0.624,0.812}
\definecolor{plum}{rgb}{0.678,0.498,0.659}
\definecolor{scarletred}{rgb}{0.937,0.161,0.161}
\definecolor{myblue}{rgb}{0.192,0.510,0.729}
\definecolor{mygreen}{rgb}{0.173,0.627,0.173}
\definecolor{myorange}{rgb}{1.000,0.498,0.055}
\definecolor{myred}{rgb}{0.839,0.153,0.157}

\usepackage{amsmath,amsfonts,amssymb,amsthm,bm}

\newcommand{\etal}{et al.}

\newcommand{\err}{\varepsilon}
\newcommand{\emperr}{\widehat{\varepsilon}}
\newcommand{\emp}{\widehat}
\newcommand{\indicator}{\boldsymbol{1}}
\newcommand{\cond}{\,|\,}

\newcommand{\newterm}[1]{{\textit{#1}}}

\def\Tableref#1{Table~\ref{#1}}

\def\Figref#1{Figure~\ref{#1}}

\def\secref#1{section~\ref{#1}}

\def\eqref#1{equation~(\ref{#1})}

\def\plaineqref#1{(\ref{#1})}  

\def\1{\bm{1}}

\def\rd{{\textnormal{d}}}

\def\rx{{\textnormal{x}}}
\def\ry{{\textnormal{y}}}
\def\rz{{\textnormal{z}}}

\def\ermX{{\textnormal{X}}}
\def\ermY{{\textnormal{Y}}}

\DeclareMathAlphabet{\mathsfit}{\encodingdefault}{\sfdefault}{m}{sl}
\SetMathAlphabet{\mathsfit}{bold}{\encodingdefault}{\sfdefault}{bx}{n}

\def\gD{{\mathcal{D}}}

\def\gF{{\mathcal{F}}}

\def\gH{{\mathcal{H}}}

\def\gL{{\mathcal{L}}}

\def\gS{{\mathcal{S}}}
\def\gT{{\mathcal{T}}}

\def\gX{{\mathcal{X}}}
\def\gY{{\mathcal{Y}}}
\def\gZ{{\mathcal{Z}}}

\def\sR{{\mathbb{R}}}

\newcommand{\E}{\mathbb{E}}

\newcommand{\softmax}{\mathrm{softmax}}

\newcommand{\normltwo}{\ell^2}

\DeclareMathOperator*{\argmax}{arg\,max}

\newtheorem{theorem}{Theorem}
\newtheorem{lemma}{Lemma}

\def\bs#1{\boldsymbol{#1}}

\def\ut#1{\underline{#1}}

\title{Tackling unsupervised multi-source domain adaptation with optimism and consistency}

\author{%
  Diogo Pernes and Jaime S. Cardoso \\
  INESC TEC and University of Porto\\
  Porto, Portugal\\
  \texttt{diogo.pernes@inesctec.pt, jaime.cardoso@inesctec.pt} \\
}

\begin{document}

\maketitle

\begin{abstract}
   It has been known for a while that the problem of multi-source domain adaptation can be regarded as a single source domain adaptation task where the source domain corresponds to a mixture of the original source domains. Nonetheless, how to adjust the mixture distribution weights remains an open question. Moreover, most existing work on this topic focuses only on minimizing the error on the source domains and achieving domain-invariant representations, which is insufficient to ensure low error on the target domain. In this work, we present a novel framework that addresses both problems and beats the current state of the art by using a mildly optimistic objective function and consistency regularization on the target samples.
\end{abstract}

\section{Introduction}
\label{sec:intro}
Supervised training of deep neural networks has achieved outstanding results on multiple learning tasks greatly due to the availability of large and rich annotated datasets. Unfortunately, annotating such large-scale datasets is often prohibitively time-consuming and expensive. Furthermore, in many practical cases, it is not possible to collect annotated data with the same characteristics as the test data, and, as a result, training and test data are drawn from distinct underlying distributions (or \newterm{domains}). As a consequence, the model performance tends to decrease significantly on the test data. The goal of \newterm{domain adaptation} (DA) algorithms is to minimize this gap by finding transferable knowledge from the source to the target domain. Sometimes, it is assumed that a small portion of labeled target data are available at training time -- a setting that is known as \newterm{semi-supervised} DA~\cite{Daume2010, Donahue2013, Kumar2010, Saito2019, Yao2015}. In this work, we focus on the more challenging scenario, where no labeled target data are available for training -- known as \newterm{unsupervised} DA~\cite{Baktashmotlagh2013, Ganin2015, Kang2019, Long2016, Zhao2018}. The DA problem, in its semi-supervised and unsupervised variants, has received increased attention in recent years, both from theoretical~\cite{BenDavid2010, BenDavid2007, Blitzer2008, Cortes2014, Gopalan2013, Hoffman2018, Zhao2019} and algorithmic perspectives~\cite{Ajakan2014, Becker2013, Fernando2013, Jhuo2012, Long2015, Louizos2015, Sun2016, Tzeng2017}. 

In many situations, the annotated training data may consist of a combination of distinct datasets, some of which may be closer or further away from the target data. Finding nontrivial ways of combining multiple datasets to approximate the target distribution and extracting relevant knowledge from such combination is the purpose of multi-source DA algorithms~\cite{Kim2017, Guo2018, Hoffman2018, Mansour2009, Sebag2019, Zhang2015, Zhao2018} and is also the focus of this work. Hoffman~\etal~\cite{Hoffman2018} focus on learning a convex combination rule to combine the classifiers of each source domain into a single classifier for the target domain. Zhao~\etal~\cite{Zhao2018} optimize the worst-case adaptation scenario among all source domains on each training iteration. Kim~\etal~\cite{Kim2017} and Guo~\etal~\cite{Guo2018} learn example-to-domain relations to weigh each source domain differently for each target example. Our approach is comparable to those of Hoffman~\etal~\cite{Hoffman2018} and Zhao~\etal~\cite{Zhao2018} in the sense that we learn a global alignment from target to source domains, but we follow a different principle: we treat the weights for each source domain as one further parameter that can be exploited to minimize the loss function and learned jointly with the remaining model parameters.

To make the DA task possible, it is often assumed that some properties are invariant across domains~\cite{Gong2016, Iyer2004, Lipton2018, Long2013, Storkey2009, Zhang2015, Zhang2013, Zhang2020}. The scenario where the marginal distribution of features changes but the conditional of labels given features is the same across domains is known as the \newterm{covariate shift} setting. It has been explored extensively either by reweighting samples from the source domain to match the target distribution~\cite{Cortes2010, Shimodaira2000, Sugiyama2008} or, more recently, by learning domain-invariant feature representations~\cite{Ajakan2014, Ferreira2019, Ganin2015, Guo2018, Pei2018, Sebag2019}. The ability of deep neural networks to learn rich feature representations and the recent surge of adversarial learning techniques have led to numerous approaches that resort to an adversarial objective to learn those representations. The adversary is usually implemented with a domain discriminator network~\cite{Ganin2015}, which aims to discriminate samples between source and target domains, and is jointly trained with the feature extractor and task classifier through a minimax game. Other models resort to the minimization of  distribution dissimilarity metrics between the target and source feature distributions~\cite{Ferreira2019, Guo2018}. It is known, however, that if the learned features violate the covariate shift assumption, domain-invariant features shall deteriorate the generalization performance of the model on the target domain~\cite{Zhao2019} -- a problem that we refer to as \newterm{the curse of domain-invariant representations}. Addressing this issue in the unsupervised setting is challenging, although some strategies have been proposed. Pei~\etal~\cite{Pei2018} use a domain discriminator per class and the probability output of the task classifier as attention weights for the discriminator. Sebag~\etal~\cite{Sebag2019} introduce a loss term that repulses unlabeled examples from the labeled ones whenever the classifier has low confidence on classifying the unlabeled examples. Thus, both methods depend on the classifier to make correct predictions on the target samples early in the training. Here, we avoid this issue by employing a consistency loss, together with a minimum confidence threshold, that enforces agreement on the class predictions for original and augmented target samples.

Our contributions are summarized as follows: i) we present a corollary of the theoretical results from~\cite{BenDavid2010} which motivates our methodology; ii) we present our multi-source adversarial model which learns the distribution weights for each source domain jointly with all remaining parameters and following a theoretically-grounded mildly optimistic approach; iii) we discuss how a simple consistency regularization technique may help avoid the curse of domain-invariant representations; iv) we conduct extensive experiments on benchmark datasets that confirm the effectiveness of the proposed methodologies.

\section{Background and motivation}
\label{sec:background}

We first introduce the problem of unsupervised multi-source DA and review the theoretical model for DA from Ben-David~\etal~\cite{BenDavid2010, BenDavid2007}.

\subsection{Problem statement}
\label{sec:problem_statement}
In the unsupervised multi-source DA setting, the learning algorithm has access to $M$ annotated datasets $S_j = \{(x_i, y_i)\}_{i=1}^{n_j} \in (\gX \times \gY)^{n_j}$, $j \in \{1,2,...,M\}$, sampled i.i.d. from $M$ joint  distributions of samples and labels, each one corresponding to a source domain $\gS_j$. We further assume to have access to a set of unlabeled examples $T=\{x_i\}_{i=1}^n \in \gX^n$, sampled i.i.d. from the marginal distribution of samples in the target domain $\gT$. In this work, we focus on classification problems, so the set $\gY$ is discrete. The ground-truth labels of the examples in $\gT$ are known to be in $\gY$, i.e. every class in the target domain is represented in the source domains. Loosely speaking, the goal of unsupervised multi-source DA is to exploit the data $S_1, ..., S_M$, and $T$ to learn a classifier with good generalization performance on the target domain.

\subsection{Notation and useful definitions}
\label{sec:notation}
Formally, a domain is a triplet $(\gX, \gY, \gD_{\ermX, \ermY})$, where $\gX$ is the input space, $\gY$ is the set of labels and $\gD_{\ermX, \ermY}$ is a joint distribution over $\gX \times \gY$ with marginals $\gD_\ermX = \int_\gY \gD_{\ermX, \ermY}$ and $\gD_\ermY = \int_\gX \gD_{\ermX, \ermY}$. In the DA setting, $\gX$ and $\gY$ are shared across all source and target domains. With some abuse of notation, from now on we denote a domain and its marginal $\gD_\ermX$ using the same symbol $\gD$. The distribution of samples in $\gD$ after a feature transformation $g:\gX \mapsto \gZ$ is denoted by $\gD^g$. For a given $\alpha$ in the simplex $\Delta$, we define $\gS_\alpha$ as the $\alpha$-weighted mixture of source domains, i.e. $\gS_\alpha = \sum_{j=1}^M \alpha_j \gS_j$. 

As is common in the DA literature, we shall restrain the theoretical analysis to binary classification. Under this setting, each domain $\gD$ is associated with a labeling function $f_\gD:\gX \mapsto [0, 1]$ that corresponds to the Bayes optimal classifier in $\gD$, $f_\gD(x) = \Pr_\gD(y=1 \cond x)$. A \newterm{hypothesis} is any function $h: \gX \mapsto \{0, 1\}$ and a set $\gH$ of hypotheses is called a \newterm{hypothesis class}. The \newterm{risk} of a hypothesis $h$ w.r.t. the domain $\gD$ is defined as $\err_\gD(h) = \E_\gD\left(|h(x) - f_\gD(x)|\right)$. Its empirical estimation is denoted by $\emperr_\gD(h)$. We further restrain to the deterministic case, where $f_\gD:\gX \mapsto \{0, 1\}$ and the label $y \in \{0, 1\}$ associated with a sample $x \in \gX$ is uniquely determined by the rule $y=f_\gD(x)$, and thus  $\err_\gD(h) = \Pr_\gD\left(h(x) \neq f_\gD(x)\right)$.

Given two distributions $\gD$ and $\gD'$ over $\gX$, the \newterm{$\gH$-divergence} \cite{BenDavid2010}, defined below, provides a measure of the distance between the distributions according to a fixed hypothesis class $\gH$:
\begin{equation}
\label{eq:h_div}
    d_{\gH}(\gD, \gD') = 2 \sup_{h \in \gH} |\Pr_{\gD}(\indicator(h)) - \Pr_{\gD'}(\indicator(h))|,
\end{equation}
where $\indicator(h) = \{x \in \gX: h(x) = 1\}$. The $\gH$-divergence is a particularly convenient metric since it can be estimated empirically from a finite number of samples whenever $\gH$ has a finite VC-dimension. This empirical estimation is denoted by $\emp{d}_\gH(\gD, \gD')$. A very useful insight is the fact that, under a weak assumption on $\gH$, computing $\emp{d}_\gH(\gD, \gD')$ is equivalent to finding the optimal classifier in $\gH$ on the task of discriminating between samples of $\gD$ and $\gD'$~\cite{BenDavid2010, BenDavid2007}. 

For a hypothesis class $\gH$, we may define the \newterm{symmetric difference hypothesis class $\gH \Delta \gH$} as:
\begin{equation}
\label{eq:h_delta_h}
    \gH \Delta \gH = \{l: l(x) = h(x) \oplus h'(x), \; h, h' \in \gH\},
\end{equation}
where $\oplus$ denotes the "exclusive or" (xor) operation. Combining this definition with \eqref{eq:h_div}, the definition of $\gH \Delta \gH$-divergence, which plays a major role in the next subsection, follows immediately. 

\subsection{An upper bound on the target risk}

Intuitively, if the source and target domains are sufficiently similar, a classifier trained to reach low empirical error on the source domain will likely achieve a low error on the target domain too. When multiple source domains are available, a combination of their respective data yields the optimal strategy. The following bound, which is a corollary of the results in~\cite{BenDavid2010}, formalizes these ideas and enlightens some properties that will be exploited by our approach. For completeness, the proof is provided in the appendix~(\secref{sec:thm_proof}).

\begin{theorem}
\label{thm:target_risk_bound}
Let $\gH$ be a hypothesis class with VC-dimension $d$. Consider an unlabeled set of $n$ samples drawn from the target domain $\gT$ and, for each $j \in \{1,2,...,M\}$, a labeled set of $n/M$ samples drawn from the source domain $\gS_j$. Then, for any $h \in \gH$ and any $\alpha \in \Delta$, with probability at least $1-\delta$ over the choice of samples,
\begin{equation}
    \label{eq:bound}
    \err_{\gT}(h) \leq \sum_{j=1}^M \alpha_j\emperr_{\gS_j}(h) + \frac{1}{2} \emp{d}_{\gH \Delta \gH}(\gS_\alpha, \gT) + \lambda_\alpha + B_\alpha(\delta) + V(\delta),
\end{equation}
where
\begin{equation}
    B_\alpha(\delta) = 2\sqrt{\frac{M\left(2d \log(2(n+1)) + \log\left(\frac{8}{\delta}\right)\right)}{n}\sum_{j=1}^M \alpha_j^2},
\end{equation}
\begin{minipage}{0.475\linewidth}
  \begin{equation}
    \lambda_\alpha = \min_{h \in \gH}\, \sum_{j=1}^M \alpha_j\err_{\gS_j}(h) + \err_{\gT}(h),
  \end{equation}
\end{minipage}\hfill
\begin{minipage}{0.475\linewidth}
  \begin{equation}
      V(\delta) = 2 \sqrt{\frac{2d\log(2n) + \log\left(\frac{4}{\delta}\right)}{n}}.
  \end{equation}
\end{minipage}
\end{theorem}
This bound is structurally very similar to the one presented in~\cite{Zhao2018}, with two slight differences: i) we work with the (empirical) $\gH \Delta \gH$-divergence between the target and the mixture of source domains directly, instead of upper-bounding it with the convex combination of (empirical) $\gH \Delta \gH$-divergences between the target and each source domain; and ii) we show the dependency of the bound on the quantity $B_\alpha(\delta)$, whose interpretation is given in the following discussion.

\subsubsection{The curse of domain-invariant representations}
\label{sec:domain_invariance}
If the optimal $\alpha$ was known, a learning algorithm for DA could, in principle, be trained to minimize the first two terms in the bound. For this purpose, it should find a function $g:\gX \mapsto \gZ$ in a set of feature transformations $\gF$ and a classifier $h: \gZ \mapsto \gY$ in $\gH$. The first term in the bound is minimized by training the classifier $h$ on the desired task, using the labeled data from all source domains. The second term is minimized by finding a feature transformation $g$ such that the induced distributions $\gT^g$ and $\gS_\alpha^g$ are similar. This is the main idea exploited by adversarial-based DA algorithms, which use an adversarial objective to match source and target distributions in a latent feature space. The problem with this approach is that it completely overlooks the role of the third term, $\lambda_\alpha$, which corresponds to the minimum possible combined risk of a classifier in $\gH$ on the source and target domains. A recent work~\cite{Zhao2019} shows constructively that a low error on the source domain and domain-invariant features are insufficient to ensure low target risk and may actually have the opposite effect, by increasing $\lambda_\alpha$. Sufficiency is established only under the covariate shift assumption, where the conditional distributions of labels $y \in \gY$ given features $z \in \gZ$ are the same across source and target domains and only the marginal distributions of features differ. Since, in the unsupervised DA setting, target labels are not available, imposing or verifying covariate shift is not possible. Moreover, whenever the marginal distributions of labels differ (target shift), a strong enforcement of domain-invariant feature representations necessarily hurts the covariate shift assumption, by imposing the marginals on features to coincide. For this reason, training adversarial-based DA algorithms for many iterations generally yields worse performance~\cite{Zhao2019}. In this work, we show empirically that learning a more robust feature transformation will generally help mitigate this issue.

\subsubsection{Choosing the combination of source domains}
\label{sec:choose_alpha}
Looking at the first two terms of the learning bound in Theorem~\ref{thm:target_risk_bound} suggests that $\alpha$ could be chosen in an optimistic way, by selecting the source domain in which the sum of the risk of the classifier and the $\gH \Delta \gH$-divergence w.r.t. the target reaches the lowest value. However, the term $B_\alpha(\delta)$ is proportional to the $\normltwo$-norm of $\alpha$, which is maximum when $\alpha$ is one-hot and minimum when $\alpha_j = 1/M$, $\forall\, j$. This has an intuitive explanation: choosing a sparse $\alpha$ implies discarding data from the source domains whose component is zero, so the classifier is trained with intrinsically less data and its error tends to increase. This discussion naturally leads to the following formal objective:
\begin{equation}
    \label{eq:formal_obj}
    \min_{\alpha \in \Delta, \, h\in \gH, \, g \in \gF}\; \sum_{j=1}^M \alpha_j\emperr_{\gS_j}(h \circ g) + \frac{1}{2} \emp{d}_{\gH \Delta \gH}(\gS_\alpha^g, \gT^g) + \mu ||\alpha||_2,
\end{equation}
where $\mu = \mu(\delta) > 0$. This objective contrasts with the idea in ~\cite{Zhao2018}, where the authors essentially choose to minimize the loss for the hardest source domain at each iteration of the learning algorithm, which is a much more pessimistic approach than ours. In either case, though, the minimum combined risk $\lambda_\alpha$ is uncontrolled and, as discussed in \secref{sec:domain_invariance}, this is an issue that must be addressed.

\section{Methodology}
\label{sec:method}

Given the discussion in the previous section, we now explain our method in detail. It basically consists of two major approaches, explained throughout this section.

\subsection{Domain adaptation from a dynamic mixture of sources}

We first address the problem of casting the objective~\plaineqref{eq:formal_obj} into a computationally treatable surrogate. As in many recent works in DA, we resort to neural networks to implement $g$ and $h$ and the empirical $\gH \Delta \gH$-divergence is implemented with a domain discriminator network, which aims to distinguish between samples of $\gS_\alpha^g$ and $\gT^g$.\footnote[1]{Theoretically, some care should be taken while designing the architecture of the discriminator to make sure that it parameterizes hypotheses in $\gH \Delta \gH$, but in practice this constraint is dropped. There are similar bounds using the $\gH$-divergence instead~\cite{Sebag2019}.} Taking the previous considerations and replacing the intractable empirical risks by the usual classification losses, objective~\plaineqref{eq:formal_obj} is cast as:
\begin{equation}
    \min_{\alpha \in \Delta, \theta_g, \theta_h} \max_{\theta_d} \; \Big\{ \gL(\alpha, \theta_g, \theta_h, \theta_d) = \gL_\text{class}(\alpha, \theta_g, \theta_h) - \mu_d \gL_\text{disc}(\alpha, \theta_g, \theta_d) + \mu_s ||\alpha||_2^2 \Big\},
\end{equation}
where 
\begin{align}
    \gL_\text{class}(\alpha, \theta_g, \theta_h) &= -\sum_{j=1}^M \alpha_j\E_{\rz \sim \gS_j^g}\left[\log p(\ry=f_{\gS_j^g}(\rz) \cond \rz, \theta_h)\right], \\
    \gL_\text{disc}(\alpha, \theta_g, \theta_d) &= -\E_{\rz \sim \gS_{\alpha}^g} \left[\log p(\rd=0 \cond \rz, \theta_d) \right] - \E_{\rz \sim \gT^g} \left[\log p(\rd=1 \cond \rz, \theta_d) \right].
\end{align}
Here, $\theta_h$ and $\theta_d$ are the parameters of the classifier and domain discriminator networks, respectively, $\mu_d, \mu_s > 0$ are hyperparameters, $\ry \in \gY$ is the categorical r.v. associated with the class label, $f_{\gS_j^g}: \gZ \mapsto \gY$ is the true (multiclass) labeling function for the $j$-th source domain, $\rd$ is a Bernoulli r.v. that discriminates source and target domains and the function $g = g(\cdot, \theta_g): \gX \mapsto \gZ$ is a neural network, parameterized by $\theta_g$, mapping input samples to features. By linearity of expectations and using the fact that $\rz = g(\rx, \theta_g)$, for input samples $\rx \in \gX$, we have:
\begin{align}
    \label{eq:loss_class}
    \gL_\text{class}(\alpha, \theta_g, \theta_h) &= -\sum_{j=1}^M \alpha_j\E_{\rx \sim \gS_j}\left[\log p(\ry=f_{\gS_j}(\rx) \cond \rx, \theta_g, \theta_h)\right], \\
    \label{eq:loss_disc}
    \gL_\text{disc}(\alpha, \theta_g, \theta_d) &= -\sum_{j=1}^M \alpha_j\E_{\rx \sim \gS_j} \left[\log p(\rd=0 \cond \rx, \theta_g, \theta_d) \right] - \E_{\rx \sim \gT} \left[\log p(\rd=1 \cond \rx, \theta_g, \theta_d) \right].
\end{align}
In order to satisfy the constraint $\alpha \in \Delta$, we reparameterize the model using $\alpha = \softmax(\beta)$, for an unconstrained parameter $\beta \in \sR^M$. Finally, using the data presented in \secref{sec:problem_statement}, we may compute empirical estimates of losses \plaineqref{eq:loss_class} and \plaineqref{eq:loss_disc}:
\begin{align}
    \label{eq:emp_loss_class}
    \gL_\text{class}(\beta, \theta_g, \theta_h) \approx& -\frac{1}{m}\sum_{j=1}^M \frac{\exp(\beta_j)}{\sum_{j'} \exp(\beta_{j'})} \sum_{(x,y) \in S_j^{(m)}} \log p(y \cond x, \theta_g, \theta_h), \\
    \label{eq:emp_loss_disc}
    \gL_\text{disc}(\beta, \theta_g, \theta_d) \approx& -\frac{1}{m}\sum_{j=1}^M \frac{\exp(\beta_j)}{\sum_{j'} \exp(\beta_{j'})} \sum_{(x,y) \in S_j^{(m)}} \log p(\rd=0 \cond x, \theta_g, \theta_d) \\
    &-\frac{1}{m}\sum_{x \in T^{(m)}} \log p(\rd=1 \cond x, \theta_g, \theta_d), \nonumber
\end{align}
where $S_j^{(m)}$ and $T^{(m)}$ are mini-batches of $m$ examples each from $S_j$ and $T$, respectively.

It is instructive to compare our approach with \cite{Zhao2018}, as that is the most similar to ours. The bound in \eqref{eq:bound} uses one single $\gH \Delta \gH$-divergence and, therefore, our model comprises one single domain discriminator, which aims to distinguish between the target and the $\alpha$-weighted mixture of source domains. In \cite{Zhao2018}, they use $M$ discriminator networks, i.e. one per source domain. More importantly, regarding the choice of $\alpha$, we treat it as one further parameter that can be optimized to minimize the loss. To avoid the objective to collapse into the easiest source domain, as explained in \secref{sec:choose_alpha}, we include an extra term that penalizes sparse $\alpha$'s. Unlike us, they do not include the sparsity penalization term and choose $\alpha$ to minimize the worst case scenario, by assigning a greater weight to the maximum loss among the $M$ source domains on each training iteration. 

\subsection{Consistency regularization on the target domain}
\label{sec:consistency}

Consistency regularization is a key component of many state of the art algorithms in semi-supervised learning. The basic idea is simple: an unlabeled sample and a slightly perturbed version of it should share the same label. Here, this approach is exploited as a sensible heuristic to avoid that domain-invariant features end up hurting the generalization performance of the model on the target domain, as discussed in \secref{sec:domain_invariance}.

\begin{figure}[h]
\centering
\fbox{\includegraphics[width=0.65\linewidth]{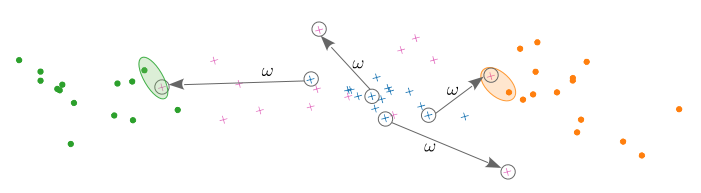}}
\caption{Toy illustration of the desired effect of the consistency regularization, where images are represented as lying on a 2-D space. Green and orange circles represent (labeled) samples from two distinct source domains; blue and purple x-markers represent original and augmented (unlabeled) target samples, respectively. Colored ellipses enclose pairs of augmented and source samples that are close to each other and therefore are likely to share the same label.}
\label{fig:consistency}
\end{figure}

Specifically, consider a target sample $x \in T$ and a parametric transformation $\omega: \gX \times \Xi \mapsto \gX$, where $\Xi$ is the space of parameters for the transformation. Further assume that $\omega$ is label-preserving, i.e. $f_\gT(\omega(x, \xi)) = f_\gT(x)$, $\forall\, x \in \gX, \, \xi \in \Xi$. If $\omega$ is rich and strong enough, it should spread the transformed target samples over multiple regions of low density under the target distribution, i.e. it will produce new domains. Then, by enforcing agreement on the predictions of original and transformed target samples, we are encouraging the model to learn to extract features that generalize well across domains. Moreover, some of the augmented samples may fall within regions of higher density under the distribution of the source domains. If this hypothesis holds, by enforcing agreement on the predictions of original and transformed target samples and low classification error on the source domains, we are indirectly promoting low error on the target domain. \Figref{fig:consistency} illustrates this idea.

The particular type of consistency regularization we adopt here is FixMatch~\cite{Sohn2020}, due to its simplicity and good performance. This approach involves using the predicted class of original (or weakly-transformed) samples as pseudo-labels for the (strongly-)transformed samples and, in our setting, is translated into the following loss function:
\begin{equation}
    \gL_{\text{cons}}(\theta_g, \theta_h) = -\frac{1}{m}\sum_{x \in T^{(m)}} \indicator\left(\max_{\ry \in \gY}\, p(\ry \cond x, \theta_g, \theta_h) > \tau \right) \log p(\tilde{y} \cond \omega(x, \xi), \theta_g, \theta_h),
\end{equation}
where $\tilde{y} = \argmax_{\ry \in \gY} p(\ry \cond x, \theta_g, \theta_h)$ is the pseudo-label, $\xi$ is chosen randomly for each $x$, and $\tau \geq 0$ is a hyperparameter defining the minimum confidence threshold for the loss to be applied. This threshold prevents the loss to be applied too early in the training process and, in our setting, may discard augmented samples that fall too far from the source distributions. The overall objective is then written as follows:
\begin{equation}
    \min_{\alpha \in \Delta, \theta_g, \theta_h}\max_{\theta_d} \Big\{ \gL = \gL_\text{class}(\alpha, \theta_g, \theta_h) + \mu_c \gL_\text{cons}(\theta_g, \theta_h) - \mu_d \gL_\text{disc}(\alpha, \theta_g, \theta_d) + \mu_s ||\alpha||_2^2 \Big\}, 
\end{equation}
where $\mu_c > 0$ controls the relative weight of the consistency loss. A full schematic of our model is provided in the appendix (\secref{sec:model_overview}).

It remains to discuss how to build the transformation $\omega$ in such a way that it is strong and diverse enough while satisfying the constraint of being label-preserving. This is essentially an application-dependent problem, though. For vision problems, there are multiple simple label-preserving transformations (e.g. translation, rotation, sharpness enhancement, etc.) that can be applied in a pipeline and, as consequence, the transformed image ends up being a strongly distorted version of the original one. This is the idea followed, for instance, in RandAugment~\cite{Cubuk2019}, which we use here. RandAugment receives as input parameters the magnitude and the number of transformations to be applied and randomly chooses the transformations to apply to each sample in the mini-batch. Here, as in~\cite{Sohn2020}, we choose the number and magnitude of the transformations uniformly at random for each mini-batch.

For generic, non-vision problems, adding random noise sampled from some known distribution (e.g. Gaussian) is a trivial realization of $\omega$. Another possibility is to apply a strong dropout~\cite{Srivastava2014} transformation at the input and inner layers of the neural network. Although such transformation is not necessarily label-preserving, dropout is known to be a successful regularization technique when applied at fully connected layers. We employ this idea in one of the experiments conducted here.

\section{Experiments}
\label{sec:experiments}

\subsection{Experimental setting}
We now conduct several experiments using standard benchmark datasets for multi-source DA. We follow established evaluation protocols for every dataset and, as far as possible, we use the same network architecture for our model and baselines. Notably, comparing with MDAN~\cite{Zhao2018}, our single domain discriminator has the same architecture as each of the $M$ domain discriminators in their model. Consequently, our model has less trainable parameters than theirs. Our parameter $\beta$ is initialized uniformly at random in $[0,1]^M$, so the resulting $\alpha$ initially weighs all source domains roughly equally, but it may become sparse as training evolves. We illustrate this behavior in the appendix~(\secref{sec:alpha_evol}). Hyperparameter tuning was performed through cross-validation over source domains and a hyperparameter sensitivity analysis is conducted in the appendix~(\secref{sec:hyperparam}). Further details about the experiments, including the label distributions on each domain, network architectures, image transformations, and search ranges for each hyperparameter, are also provided in the appendix (sections~\ref{sec:label_dist},~\ref{sec:arch},~\ref{sec:cross_val},~and~\ref{sec:img_transf}). The PyTorch-based implementation of our model is publicly available.\footnote{\url{https://github.com/dpernes/modafm}} To facilitate the presentation, from now on we refer to our model as MODA-FM (Multi-source mildly Optimistic Domain Adaptation with FixMatch regularization). 

\paragraph{Baselines} DANN-SS~\cite{Ganin2015}: a single-source DA model, where the best results among all source domains are reported. DANN-MS: the same model as before trained on the combined data from all source domains. MADA~\cite{Pei2018}: a state of the art model for single-source DA which tries to align conditional distributions by using one domain discriminator per class (results extracted from~\cite{Pei2018}, we report the best accuracy among all source domains). MDAN~\cite{Zhao2018}: a state of the art model for multi-source DA, widely described before. MoE~\cite{Guo2018}: a state of the art model for multi-source DA that uses one classifier per source domain whose predictions are weighted in an example-dependent way. Fully supervised: fully supervised model trained on the target data, to provide an empirical upper bound on the performance of the DA task. MODA: our model without consistency regularization (i.e., $\mu_c = 0$). FM: our model without domain discriminator, trained on the naively combined data from all source domains and using FixMatch consistency regularization as described in \secref{sec:consistency}.

\paragraph{Digits classification} In this experiment, the task is digit classification using 4 datasets: MNIST~\cite{LeCun1998}, MNIST-M~\cite{Ganin2015}, SVHN~\cite{Netzer2011}, and SynthDigits~\cite{Ganin2015}. We take each of the first three datasets as the target in turn, and use the remaining as source domains. The number of training images chosen randomly from each domain, including the target, is 20k. The evaluation is performed in the non-transductive setting, i.e. no target data used during training are used for evaluation. The results are in \Tableref{tab:digits_office_acc}.

\begin{table}
\centering
\caption{Average accuracy $\pm$ standard deviation (\%) over 5 independent runs on digits and objects classification. The domain on each column corresponds to the target.}
\small
\label{tab:digits_office_acc}
\begin{tabular}{l|ccc|ccc}
                          & MNIST                        & MNIST-M                       & SVHN                         & Amazon                        & DSLR                              & Webcam                       \\ \hline
DANN-SS~\cite{Ganin2015}  & $ 97.9 $ \tiny{$ \pm 0.4 $}  & $ 73.2 $ \tiny{$ \pm 1.6 $}   & $ 72.8 $ \tiny{$ \pm 3.3 $}  & $ 60.5 $ \tiny{$ \pm 1.4 $}   & $\bs{100.0}$ \tiny{$\pm 0.0$}     & $ 98.0$ 	\tiny{$ \pm 0.3 $} \\
DANN-MS~\cite{Ganin2015}  & $ 97.9 $ \tiny{$ \pm 1.6 $}  & $ 67.5 $ \tiny{$ \pm 1.4 $}   & $ 71.5 $ \tiny{$ \pm 1.6 $}  & $ 61.2 $ \tiny{$ \pm 1.3 $}   & $ 99.9 $ \tiny{$ \pm 0.2 $}       & $ 98.8 $ \tiny{$ \pm 0.3 $}  \\
MADA~\cite{Pei2018}       & --                        & --                             & --                           & $ 70.3 $ \tiny{$ \pm 0.3 $}   & $ 99.6 $ \tiny{$ \pm 0.1 $}       & $ 97.4 $ \tiny{$ \pm 0.1 $}  \\
MDAN~\cite{Zhao2018}      & $ 98.3 $ \tiny{$ \pm 0.2 $}  & $ 69.1 $ \tiny{$ \pm 1.2 $}   & $ 69.5 $ \tiny{$ \pm 2.8 $}  & $ 65.2 $ \tiny{$ \pm 0.4 $}   & $ 99.3 $ \tiny{$ \pm 0.2 $}       & $ 97.8 $ \tiny{$ \pm 0.5 $}  \\
MoE~\cite{Guo2018}        & $ 98.6 $ \tiny{$ \pm 0.2 $}                     & $ 69.9 $ \tiny{$ \pm 1.1 $}                      & $ 81.8 $ \tiny{$ \pm 0.8 $}                     & --                            & --                                & --                           \\ \hline
MODA         & $ 98.4 $ \tiny{$ \pm 0.2 $}  & $ 77.4 $ \tiny{$ \pm 1.6 $}   & $ 71.7 $ \tiny{$ \pm 1.5 $}  & $ 65.5 $ \tiny{$ \pm 0.5 $}   &   $ 99.9 $ \tiny{$ \pm 0.2 $}     & $ 99.0 $ \tiny{$ \pm 0.3 $}  \\
FM           & $\bs{99.2}$ \tiny{$\pm 0.1$} & $ 91.1 $ \tiny{$ \pm 0.4 $}   & $\bs{90.0}$ \tiny{$\pm 0.8$} & $ 70.3 $ \tiny{$ \pm 0.6 $}   & $ 99.7 $ \tiny{$ \pm 0.2 $}       & $\bs{99.2}$ \tiny{$\pm 0.4$} \\
MODA-FM                      & $ 98.8 $ \tiny{$ \pm 0.1 $}  & $\bs{95.4}$ \tiny{$\pm 0.4$}  & $ 89.4 $ \tiny{$ \pm 1.4 $}  & $\bs{70.7}$ \tiny{$\pm 0.9$}  & $ \bs{100.0} $ \tiny{$ \pm 0.0 $} & $ 99.1 $ \tiny{$ \pm 0.1 $}  \\ \hline
Fully supervised            & $ 98.9 $ \tiny{$ \pm 0.1 $}  & $ 96.2 $ \tiny{$ \pm 0.1 $}   & $ 90.3 $ \tiny{$ \pm 0.5 $}  & $ 88.1 $ \tiny{$ \pm 1.6 $}   & $ 99.3 $ \tiny{$ \pm 1.1 $}       & $ 99.5 $ \tiny{$ \pm 0.7 $}
\end{tabular}
\end{table}

\paragraph{Object classification on Office-31 dataset} Office-31~\cite{Saenko2010} is a standard benchmark dataset for domain adaptation. It comprises 31 object categories extracted from 3 domains: Amazon, which contains 2817 images downloaded from \url{amazon.com}, DSLR and Webcam, which contain 498 and 795 images captured with DSLR cameras and webcams, respectively, under different environments. In this experiment, we adopt the fully-transductive setting, following~\cite{Pei2018}, where all unlabeled data from the target domain are used for training (except for the fully supervised model, where we use 80\% of the data for training and the remaining for testing). All models are implemented using a pre-trained (on ImageNet) ResNet-50~\cite{He2016} as the base architecture. We take each domain as the target in turn and all remaining are used as sources. The results are presented in \Tableref{tab:digits_office_acc}.

\paragraph{Sentiment analysis on Amazon Reviews dataset} The Amazon Reviews dataset~\cite{Blitzer2007} is another multi-domain dataset widely used as a benchmark for domain adaptation. It contains binary (i.e., positive and negative) reviews on 4 types of products: books, DVDs, electronics and kitchen appliances. Here, we follow the experimental setting from~\cite{Chen2012}, where samples were pre-processed to 5k-dimensional TF-IDF feature vectors, thus word order information was not preserved. We choose 2k training samples from each domain randomly and the remaining target samples are used for testing, following the non-transductive setting. Since the data are not images, we use dropout as our (pseudo-)label-preserving transformation. Specifically, on each training iteration, we randomly choose a dropout rate (in a pre-specified range) to be applied at the input and hidden layers of the MLP and the corresponding output prediction is used as the pseudo-label $\tilde{y}$. No dropout is applied for source and non-augmented target samples. All domains are taken as the target in turn and all remaining are used as sources. The results are in~\Tableref{tab:amazon_acc}.

\begin{table}[]
\centering
\caption{Average accuracy $\pm$ standard deviation (\%) over 5 independent runs on sentiment analysis (Amazon Reviews). The domain on each column corresponds to the target.}
\label{tab:amazon_acc}
\small
\begin{tabular}{l|cccc}
                         & Books           & DVD             & Electronics     & Kitchen         \\ \hline
DANN-SS~\cite{Ganin2015} & $ 77.5 $ \tiny{$ \pm 1.0 $} & $ 78.9 $ \tiny{$ \pm 0.9 $} & $ 84.2 $ \tiny{$ \pm 0.2 $} & $ 85.8 $ \tiny{$ \pm 0.4 $} \\
DANN-MS~\cite{Ganin2015} & $ 78.9 $ \tiny{$ \pm 0.2 $} & $ 80.8 $ \tiny{$ \pm 1.0 $} & $ 84.7 $ \tiny{$ \pm 0.6 $} & $ 87.0 $ \tiny{$ \pm 0.4 $} \\
MDAN~\cite{Zhao2018}     & $ 79.1 $ \tiny{$ \pm 0.3 $} & $ 81.3 $ \tiny{$ \pm 0.8 $} & $ 84.6 $ \tiny{$ \pm 0.3 $} & $ 85.6 $ \tiny{$ \pm 1.6 $} \\
MoE~\cite{Guo2018}       & $ 80.3 $ \tiny{$ \pm 0.3 $} & $ 81.9 $ \tiny{$ \pm 0.6 $} & $ 85.2 $ \tiny{$ \pm 0.6 $} & $ 87.4 $ \tiny{$ \pm 0.5 $} \\ \hline
MODA     & $ 79.0 $ \tiny{$ \pm 0.2 $} & $ 80.7 $ \tiny{$ \pm 1.2 $} & $ 84.7 $ \tiny{$ \pm 0.3 $} & $ 86.8 $ \tiny{$ \pm 0.6 $} \\
FM         & $ 78.8 $ \tiny{$ \pm 2.0 $} & $ 81.0 $ \tiny{$ \pm 1.5 $} & $\bs{85.6}$ \tiny{$\pm 0.9 $} & $ 87.6 $ \tiny{$ \pm 0.7 $} \\
MODA-FM    & $\bs{80.9}$ \tiny{$\pm 1.1$} & $\bs{82.0}$ \tiny{$\pm 1.0$} & $ 85.3 $ \tiny{$ \pm 0.4 $} & $\bs{88.4}$ \tiny{$\pm 0.3$} \\ \hline
Fully supervised & $ 83.6 $ \tiny{$ \pm 0.4 $} & $ 83.6 $ \tiny{$ \pm 0.6 $} & $ 85.4 $ \tiny{$ \pm 0.4 $} & $ 87.8 $ \tiny{$ \pm 0.2 $}
\end{tabular}
\end{table} 

\subsection{Discussion}
Tables~\ref{tab:digits_office_acc}~and~\ref{tab:amazon_acc} show that our model outperforms the baselines in most settings and performs comparably to the fully-supervised model in many. When no consistency regularization is used (MODA), our approach exhibits a higher accuracy than DANN-SS, DANN-MS and MDAN in most cases. This observation shows that our mildly optimistic combination of source domains works better than using only the data from the best source domain (DANN-SS) or naively combining the data from all source domains (DANN-MS). It also suggests that MDAN wastes too much computational effort on optimizing itself for the hardest source domain. MADA and MoE perform better than our non-regularized model in some cases, which is not surprising since both methods try to mitigate somehow the curse of domain-invariant representations. Their advantage is, therefore, mostly noticeable when the target shift is large (e.g., SVHN and all domains in Office-31 -- see the appendix, \secref{sec:label_dist}), but the performance is still below MODA-FM.

Very significant performance gains are observed when we apply the consistency regularization, particularly in the visual datasets (digits and Office-31). These gains are more expressive in the most challenging settings, i.e. when the target data are perceptually very different from the source data (e.g., MNIST-M -- see the appendix, section~C.8) or when the target shift is large. The latter observation suggests that this regularization succeeds on mitigating the curse of domain-invariant representations, as hypothesized before. This is strongly corroborated by an experiment we present in the appendix~(\secref{sec:overtrain}), where we show that the model accuracy on the target domain keeps stably high when the model is trained for a large number of epochs. Interestingly, though, we observe that, in some settings, FM outperforms MODA-FM, although the differences are small. In these cases domain-invariant representations are slightly hurting the performance and/or the source weights $\alpha$ are sub-optimal. Finally, it is worth highlighting the positive effect provided by using dropout as the label-preserving transformation $\omega$ in the experiment with Amazon Reviews. This observation suggests that this methodology can be applied successfully to non-visual data too.

\section{Conclusion}
\label{sec:conclusion}
We have presented a novel algorithm that achieves state of the art results in unsupervised multi-source domain adaptation. In our approach, the problem is formulated as DA from a single source domain whose distribution corresponds to a mixture of the original source domains. The mixture weights are adjusted dynamically throughout the training process, according to a mildly optimistic objective. Additionally, we employ FixMatch on the target samples, a form of consistency regularization that proves to have a strong impact on the model performance and to be capable of mitigating the curse of domain invariant representations. This regularization relies on a label-preserving transformation, which is hard to construct for non-visual data. Moreover, better results could be achieved if both the label-preserving transformation and the source weights were learned to approximate the augmented target samples close to the source samples. Both problems are interesting lines for future research.

\begin{ack}
This work was funded by FCT - Funda\c{c}\~{a}o para a Ci\^{e}ncia e a Tecnologia within Ph.D. grant number SFRH/BD/129600/2017.
\end{ack}

\begin{small}
\bibliography{refs}

\begin{thebibliography}{10}

\bibitem{Ajakan2014}
H.~Ajakan, P.~Germain, H.~Larochelle, F.~Laviolette, and M.~Marchand.
\newblock Domain-adversarial neural networks.
\newblock {\em arXiv preprint arXiv:1412.4446}, 2014.

\bibitem{Baktashmotlagh2013}
M.~Baktashmotlagh, M.~T. Harandi, B.~C. Lovell, and M.~Salzmann.
\newblock Unsupervised domain adaptation by domain invariant projection.
\newblock In {\em Proceedings of the IEEE International Conference on Computer
  Vision}, pages 769--776, 2013.

\bibitem{Becker2013}
C.~J. Becker, C.~M. Christoudias, and P.~Fua.
\newblock Non-linear domain adaptation with boosting.
\newblock In {\em Advances in Neural Information Processing Systems}, pages
  485--493, 2013.

\bibitem{BenDavid2010}
S.~Ben-David, J.~Blitzer, K.~Crammer, A.~Kulesza, F.~Pereira, and J.~W.
  Vaughan.
\newblock A theory of learning from different domains.
\newblock {\em Machine learning}, 79(1-2):151--175, 2010.

\bibitem{BenDavid2007}
S.~Ben-David, J.~Blitzer, K.~Crammer, and F.~Pereira.
\newblock Analysis of representations for domain adaptation.
\newblock In {\em Advances in Neural Information Processing Systems}, pages
  137--144, 2007.

\bibitem{Blitzer2008}
J.~Blitzer, K.~Crammer, A.~Kulesza, F.~Pereira, and J.~Wortman.
\newblock Learning bounds for domain adaptation.
\newblock In {\em Advances in Neural Information Processing Systems}, pages
  129--136, 2008.

\bibitem{Blitzer2007}
J.~Blitzer, M.~Dredze, and F.~Pereira.
\newblock Biographies, bollywood, boom-boxes and blenders: Domain adaptation
  for sentiment classification.
\newblock In {\em Proceedings of the 45th annual meeting of the association of
  computational linguistics}, pages 440--447, 2007.

\bibitem{Chen2012}
M.~Chen, Z.~Xu, K.~Weinberger, and F.~Sha.
\newblock Marginalized denoising autoencoders for domain adaptation.
\newblock {\em arXiv preprint arXiv:1206.4683}, 2012.

\bibitem{Cortes2010}
C.~Cortes, Y.~Mansour, and M.~Mohri.
\newblock Learning bounds for importance weighting.
\newblock In {\em Advances in Neural Information Processing Systems}, pages
  442--450, 2010.

\bibitem{Cortes2014}
C.~Cortes and M.~Mohri.
\newblock Domain adaptation and sample bias correction theory and algorithm for
  regression.
\newblock {\em Theoretical Computer Science}, 519:103--126, 2014.

\bibitem{Cubuk2019}
E.~D. Cubuk, B.~Zoph, J.~Shlens, and Q.~V. Le.
\newblock Randaugment: Practical data augmentation with no separate search.
\newblock {\em arXiv preprint arXiv:1909.13719}, 2019.

\bibitem{Daume2010}
H.~Daum{\'e}~III, A.~Kumar, and A.~Saha.
\newblock Frustratingly easy semi-supervised domain adaptation.
\newblock In {\em Proceedings of the 2010 Workshop on Domain Adaptation for
  Natural Language Processing}, pages 53--59. Association for Computational
  Linguistics, 2010.

\bibitem{Donahue2013}
J.~Donahue, J.~Hoffman, E.~Rodner, K.~Saenko, and T.~Darrell.
\newblock Semi-supervised domain adaptation with instance constraints.
\newblock In {\em Proceedings of the IEEE Conference on Computer Vision and
  Pattern Recognition}, pages 668--675, 2013.

\bibitem{Fernando2013}
B.~Fernando, A.~Habrard, M.~Sebban, and T.~Tuytelaars.
\newblock Unsupervised visual domain adaptation using subspace alignment.
\newblock In {\em Proceedings of the IEEE International Conference on Computer
  Vision}, pages 2960--2967, 2013.

\bibitem{Ferreira2019}
P.~M. Ferreira, D.~Pernes, A.~Rebelo, and J.~S. Cardoso.
\newblock Desire: Deep signer-invariant representations for sign language
  recognition.
\newblock {\em IEEE Transactions on Systems, Man, and Cybernetics: Systems},
  2019.

\bibitem{Ganin2015}
Y.~Ganin and V.~Lempitsky.
\newblock Unsupervised domain adaptation by backpropagation.
\newblock In {\em International Conference on Machine Learning}, pages
  1180--1189, 2015.

\bibitem{Gong2016}
M.~Gong, K.~Zhang, T.~Liu, D.~Tao, C.~Glymour, and B.~Sch{\"o}lkopf.
\newblock Domain adaptation with conditional transferable components.
\newblock In {\em International Conference on Machine Learning}, pages
  2839--2848, 2016.

\bibitem{Gopalan2013}
R.~Gopalan, R.~Li, and R.~Chellappa.
\newblock Unsupervised adaptation across domain shifts by generating
  intermediate data representations.
\newblock {\em IEEE Transactions on Pattern Analysis and Machine Intelligence},
  36(11):2288--2302, 2013.

\bibitem{Guo2018}
J.~Guo, D.~Shah, and R.~Barzilay.
\newblock Multi-source domain adaptation with mixture of experts.
\newblock In {\em Proceedings of the 2018 Conference on Empirical Methods in
  Natural Language Processing}, pages 4694--4703, 2018.

\bibitem{He2016}
K.~He, X.~Zhang, S.~Ren, and J.~Sun.
\newblock Deep residual learning for image recognition.
\newblock In {\em Proceedings of the IEEE Conference on Computer Vision and
  Pattern Recognition}, pages 770--778, 2016.

\bibitem{Hoffman2018}
J.~Hoffman, M.~Mohri, and N.~Zhang.
\newblock Algorithms and theory for multiple-source adaptation.
\newblock In {\em Advances in Neural Information Processing Systems}, pages
  8246--8256, 2018.

\bibitem{Iyer2004}
A.~Iyer, S.~Nath, and S.~Sarawagi.
\newblock Maximum mean discrepancy for class ratio estimation: Convergence
  bounds and kernel selection.
\newblock In {\em International Conference on Machine Learning}, pages
  530--538, 2014.

\bibitem{Jhuo2012}
I.-H. Jhuo, D.~Liu, D.~Lee, and S.-F. Chang.
\newblock Robust visual domain adaptation with low-rank reconstruction.
\newblock In {\em 2012 IEEE Conference on Computer Vision and Pattern
  Recognition}, pages 2168--2175. IEEE, 2012.

\bibitem{Kang2019}
G.~Kang, L.~Jiang, Y.~Yang, and A.~G. Hauptmann.
\newblock Contrastive adaptation network for unsupervised domain adaptation.
\newblock In {\em Proceedings of the IEEE Conference on Computer Vision and
  Pattern Recognition}, pages 4893--4902, 2019.

\bibitem{Kim2017}
Y.-B. Kim, K.~Stratos, and D.~Kim.
\newblock Domain attention with an ensemble of experts.
\newblock In {\em Proceedings of the 55th Annual Meeting of the Association for
  Computational Linguistics (Volume 1: Long Papers)}, pages 643--653, 2017.

\bibitem{Kumar2010}
A.~Kumar, A.~Saha, and H.~Daume.
\newblock Co-regularization based semi-supervised domain adaptation.
\newblock In {\em Advances in Neural Information Processing Systems}, pages
  478--486, 2010.

\bibitem{LeCun1998}
Y.~LeCun, L.~Bottou, Y.~Bengio, and P.~Haffner.
\newblock Gradient-based learning applied to document recognition.
\newblock {\em Proceedings of the IEEE}, 86(11):2278--2324, 1998.

\bibitem{Lipton2018}
Z.~C. Lipton, Y.-X. Wang, and A.~Smola.
\newblock Detecting and correcting for label shift with black box predictors.
\newblock {\em arXiv preprint arXiv:1802.03916}, 2018.

\bibitem{Long2015}
M.~Long, Y.~Cao, J.~Wang, and M.~I. Jordan.
\newblock Learning transferable features with deep adaptation networks.
\newblock {\em arXiv preprint arXiv:1502.02791}, 2015.

\bibitem{Long2013}
M.~Long, J.~Wang, G.~Ding, J.~Sun, and P.~S. Yu.
\newblock Transfer feature learning with joint distribution adaptation.
\newblock In {\em Proceedings of the IEEE International Conference on Computer
  Vision}, pages 2200--2207, 2013.

\bibitem{Long2016}
M.~Long, H.~Zhu, J.~Wang, and M.~I. Jordan.
\newblock Unsupervised domain adaptation with residual transfer networks.
\newblock In {\em Advances in Neural Information Processing Systems}, pages
  136--144, 2016.

\bibitem{Louizos2015}
C.~Louizos, K.~Swersky, Y.~Li, M.~Welling, and R.~Zemel.
\newblock The variational fair autoencoder.
\newblock In {\em International Conference on Learning Representations}, 2016.

\bibitem{Mansour2009}
Y.~Mansour, M.~Mohri, and A.~Rostamizadeh.
\newblock Domain adaptation: Learning bounds and algorithms.
\newblock In {\em 22nd Conference on Learning Theory, COLT 2009}, 2009.

\bibitem{Netzer2011}
Y.~Netzer, T.~Wang, A.~Coates, A.~Bissacco, B.~Wu, and A.~Y. Ng.
\newblock Reading digits in natural images with unsupervised feature learning.
\newblock 2011.

\bibitem{Pei2018}
Z.~Pei, Z.~Cao, M.~Long, and J.~Wang.
\newblock Multi-adversarial domain adaptation.
\newblock In {\em Thirty-Second AAAI Conference on Artificial Intelligence},
  2018.

\bibitem{Saenko2010}
K.~Saenko, B.~Kulis, M.~Fritz, and T.~Darrell.
\newblock Adapting visual category models to new domains.
\newblock In {\em European Conference on Computer Vision}, pages 213--226.
  Springer, 2010.

\bibitem{Saito2019}
K.~Saito, D.~Kim, S.~Sclaroff, T.~Darrell, and K.~Saenko.
\newblock Semi-supervised domain adaptation via minimax entropy.
\newblock In {\em Proceedings of the IEEE International Conference on Computer
  Vision}, pages 8050--8058, 2019.

\bibitem{Sebag2019}
A.~Schoenauer-Sebag, L.~Heinrich, M.~Schoenauer, M.~Sebag, L.~Wu, and
  S.~Altschuler.
\newblock Multi-domain adversarial learning.
\newblock In {\em International Conference on Learning Representations}, 2019.

\bibitem{Shimodaira2000}
H.~Shimodaira.
\newblock Improving predictive inference under covariate shift by weighting the
  log-likelihood function.
\newblock {\em Journal of statistical planning and inference}, 90(2):227--244,
  2000.

\bibitem{Sohn2020}
K.~Sohn, D.~Berthelot, C.-L. Li, Z.~Zhang, N.~Carlini, E.~D. Cubuk, A.~Kurakin,
  H.~Zhang, and C.~Raffel.
\newblock Fixmatch: Simplifying semi-supervised learning with consistency and
  confidence.
\newblock {\em arXiv preprint arXiv:2001.07685}, 2020.

\bibitem{Srivastava2014}
N.~Srivastava, G.~Hinton, A.~Krizhevsky, I.~Sutskever, and R.~Salakhutdinov.
\newblock Dropout: a simple way to prevent neural networks from overfitting.
\newblock {\em The journal of machine learning research}, 15(1):1929--1958,
  2014.

\bibitem{Storkey2009}
A.~Storkey.
\newblock When training and test sets are different: characterizing learning
  transfer.
\newblock {\em Dataset shift in machine learning}, pages 3--28, 2009.

\bibitem{Sugiyama2008}
M.~Sugiyama, T.~Suzuki, S.~Nakajima, H.~Kashima, P.~von B{\"u}nau, and
  M.~Kawanabe.
\newblock Direct importance estimation for covariate shift adaptation.
\newblock {\em Annals of the Institute of Statistical Mathematics},
  60(4):699--746, 2008.

\bibitem{Sun2016}
B.~Sun, J.~Feng, and K.~Saenko.
\newblock Return of frustratingly easy domain adaptation.
\newblock In {\em Thirtieth AAAI Conference on Artificial Intelligence}, 2016.

\bibitem{Tzeng2017}
E.~Tzeng, J.~Hoffman, K.~Saenko, and T.~Darrell.
\newblock Adversarial discriminative domain adaptation.
\newblock In {\em Proceedings of the IEEE Conference on Computer Vision and
  Pattern Recognition}, pages 7167--7176, 2017.

\bibitem{Yao2015}
T.~Yao, Y.~Pan, C.-W. Ngo, H.~Li, and T.~Mei.
\newblock Semi-supervised domain adaptation with subspace learning for visual
  recognition.
\newblock In {\em Proceedings of the IEEE Conference on Computer Vision and
  Pattern Recognition}, pages 2142--2150, 2015.

\bibitem{Zhang2015}
K.~Zhang, M.~Gong, and B.~Sch{\"o}lkopf.
\newblock Multi-source domain adaptation: A causal view.
\newblock In {\em Twenty-ninth AAAI conference on artificial intelligence},
  2015.

\bibitem{Zhang2020}
K.~Zhang, M.~Gong, P.~Stojanov, B.~Huang, and C.~Glymour.
\newblock Domain adaptation as a problem of inference on graphical models.
\newblock {\em arXiv preprint arXiv:2002.03278}, 2020.

\bibitem{Zhang2013}
K.~Zhang, B.~Sch{\"o}lkopf, K.~Muandet, and Z.~Wang.
\newblock Domain adaptation under target and conditional shift.
\newblock In {\em International Conference on Machine Learning}, pages
  819--827, 2013.

\bibitem{Zhao2019}
H.~Zhao, R.~T. Des~Combes, K.~Zhang, and G.~Gordon.
\newblock On learning invariant representations for domain adaptation.
\newblock In {\em International Conference on Machine Learning}, pages
  7523--7532, 2019.

\bibitem{Zhao2018}
H.~Zhao, S.~Zhang, G.~Wu, J.~M. Moura, J.~P. Costeira, and G.~J. Gordon.
\newblock Adversarial multiple source domain adaptation.
\newblock In {\em Advances in Neural Information Processing Systems}, pages
  8559--8570, 2018.

\end{thebibliography}
\bibliographystyle{abbrv}
\end{small}

\appendix
\section{Proof of Theorem 1}
\label{sec:thm_proof}

The presented bound is an immediate consequence of two theorems of Ben-David~\cite{BenDavid2010} and Blitzer~\cite{Blitzer2008}, which we present here as lemmas:

\begin{lemma}
\label{lemma:bound_mixture}
(Lemma 4 in \cite{Blitzer2008}) Let $\gH$ be a hypothesis class with VC-dimension $d$. For each $j \in \{1,2,...,M\}$, consider a labeled set of $n/M$ samples drawn from the source domain $\gS_j$. For any $h \in \gH$ and any $\alpha \in \Delta$, with probability at least $1-\delta$ over the choice of samples,

\begin{equation}
    |\emperr_\alpha(h) - \err_\alpha(h)| \leq 2\sqrt{\frac{M(2d\log(2(n+1)) + \log(\frac{4}{\delta}))}{n} \sum_{j=1}^M \alpha_j^2}.
\end{equation}
\end{lemma}

\begin{lemma}
\label{lemma:bound_single_source}
(Theorem 2 in \cite{BenDavid2010}) Let $\gH$ be a hypothesis class with VC-dimension $d$. Consider $n$ unlabeled samples drawn from each of the two domains $\gS$ (source) and $\gT$ (target). Then, for every $h \in \gH$, with probability at least $1-\delta$ over the choice of samples,

\begin{equation}
    \err_\gT(h) \leq \err_\gS(h) + \frac{1}{2} \emp{d}_{\gH \Delta \gH}(\gS, \gT) + 2\sqrt{\frac{2d\log(2n) + \log(\frac{2}{\delta})}{n}} + \lambda,
\end{equation}    
where $\lambda = \min_{h \in \gH} \err_\gS(h) + \err_\gT(h)$.
\end{lemma}

Proving our result is now straightforward. We can apply Lemma~\ref{lemma:bound_single_source} taking $\gS_\alpha = \sum_{j=1}^M \alpha_j \gS_j$ as the source domain. Note that $\err_{S_\alpha}(h) = \sum_{j=1}^M \alpha_j \err_{S_j}(h)$ for any $h \in \gH$, and therefore $\lambda_\alpha = \min_{h \in \gH}\, \sum_{j=1}^M \alpha_j\err_{\gS_j}(h) + \err_{\gT}(h)$. Combining the obtained bound with Lemma~\ref{lemma:bound_mixture} through the union bound yields the desired result. $\square$

\section{Model overview}
\label{sec:model_overview}

\tikzset{every picture/.style={line width=0.75pt}} 
\begin{figure}[h!]
 \centering
 \scalebox{.69}{\input{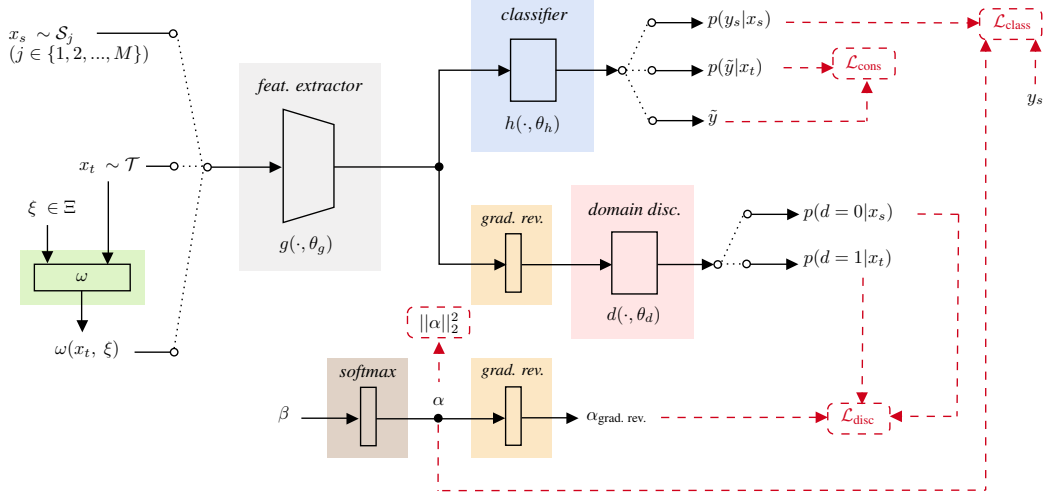}}
 \caption{Block diagram representing the proposed model (MODA-FM).}
  \label{fig:model_diagram}
\end{figure}

To enhance clarity, a detailed schematic of our model and loss terms is shown in \Figref{fig:model_diagram}. The three main blocks are the feature extractor $g(\cdot, \theta_g)$, the classifier $h(\cdot, \theta_h)$, and the domain discriminator $d(\cdot, \theta_d)$. It is worth noting the usage of gradient reversal layers~\cite{Ganin2015} at the input of the domain discriminator and before the $\alpha$ that is used in the discriminator loss $\gL_\text{disc}$. This layer is the identity function in the forward pass but reverses the gradient in the backward pass, by multiplying it by a negative constant (here, $-1$ is used). Hence, the usage of this layer allows the model to be trained using standard backpropagation and gradient descent (or any of its variants) over all model parameters.

\section{Experiments -- further results and details}

\subsection{Label distributions}
\label{sec:label_dist}
\begin{figure}
  \centering
  \begin{subfigure}[b]{0.24\textwidth}
    \centering
    \includegraphics[width=\textwidth]{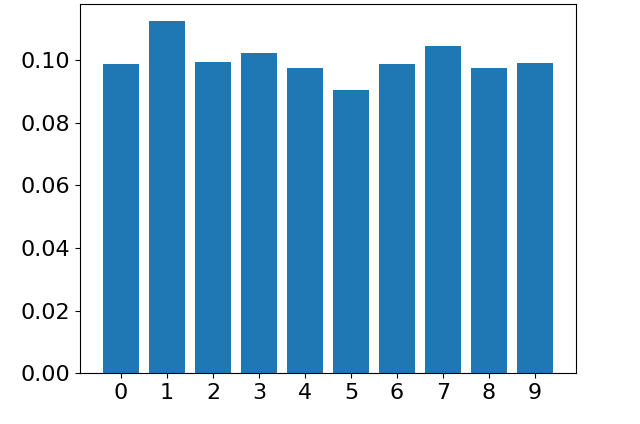}
    \caption{MNIST}
  \end{subfigure}
  \hfill
  \begin{subfigure}[b]{0.24\textwidth}  
    \centering 
    \includegraphics[width=\textwidth]{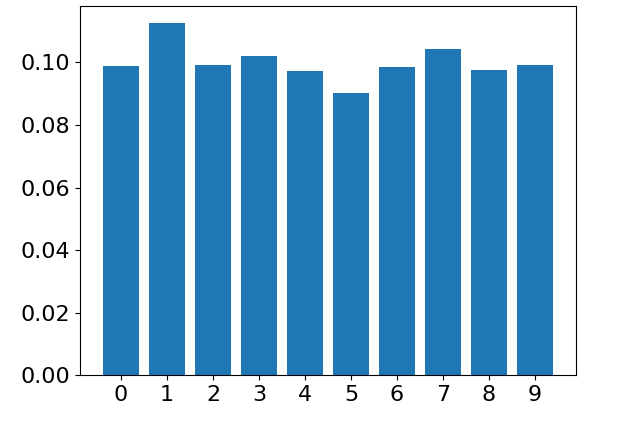}
    \caption{MNIST-M}
  \end{subfigure}
  \hfill
  \begin{subfigure}[b]{0.24\textwidth}   
    \centering 
    \includegraphics[width=\textwidth]{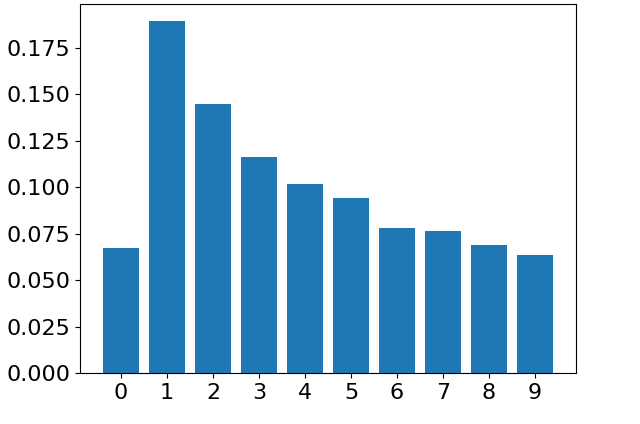}
    \caption{SVHN}
  \end{subfigure}
  \hfill
  \begin{subfigure}[b]{0.24\textwidth}   
    \centering 
    \includegraphics[width=\textwidth]{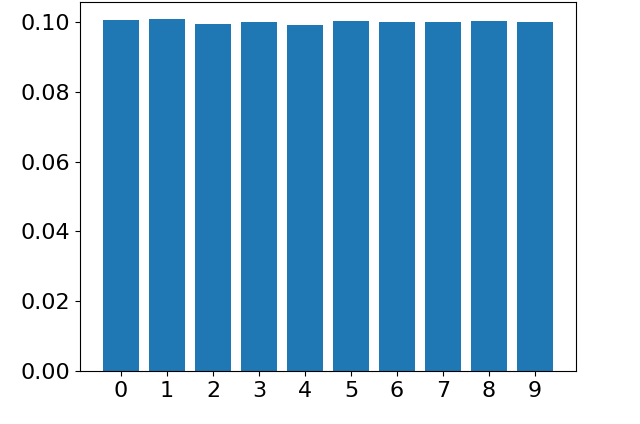}
    \caption{SynthDigits}
  \end{subfigure}
  \caption{Label distributions in the digits datasets.} 
  \label{fig:digits_dist}
\end{figure}

\begin{figure}
  \centering
  \begin{subfigure}[b]{0.35\textwidth}
    \centering
    \includegraphics[width=\textwidth]{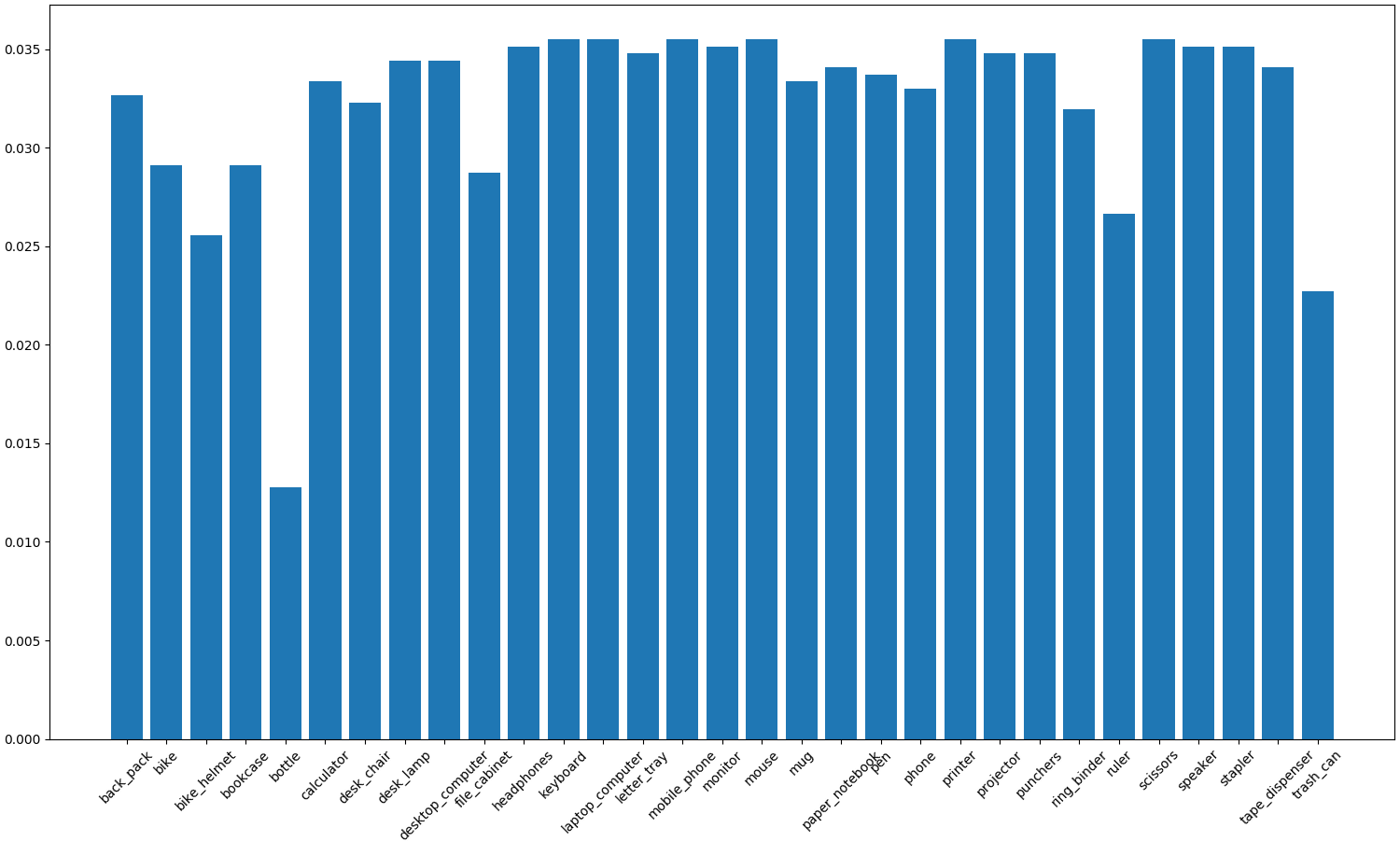}
    \caption{Amazon}
  \end{subfigure}
  \;
  \begin{subfigure}[b]{0.35\textwidth}  
    \centering 
    \includegraphics[width=\textwidth]{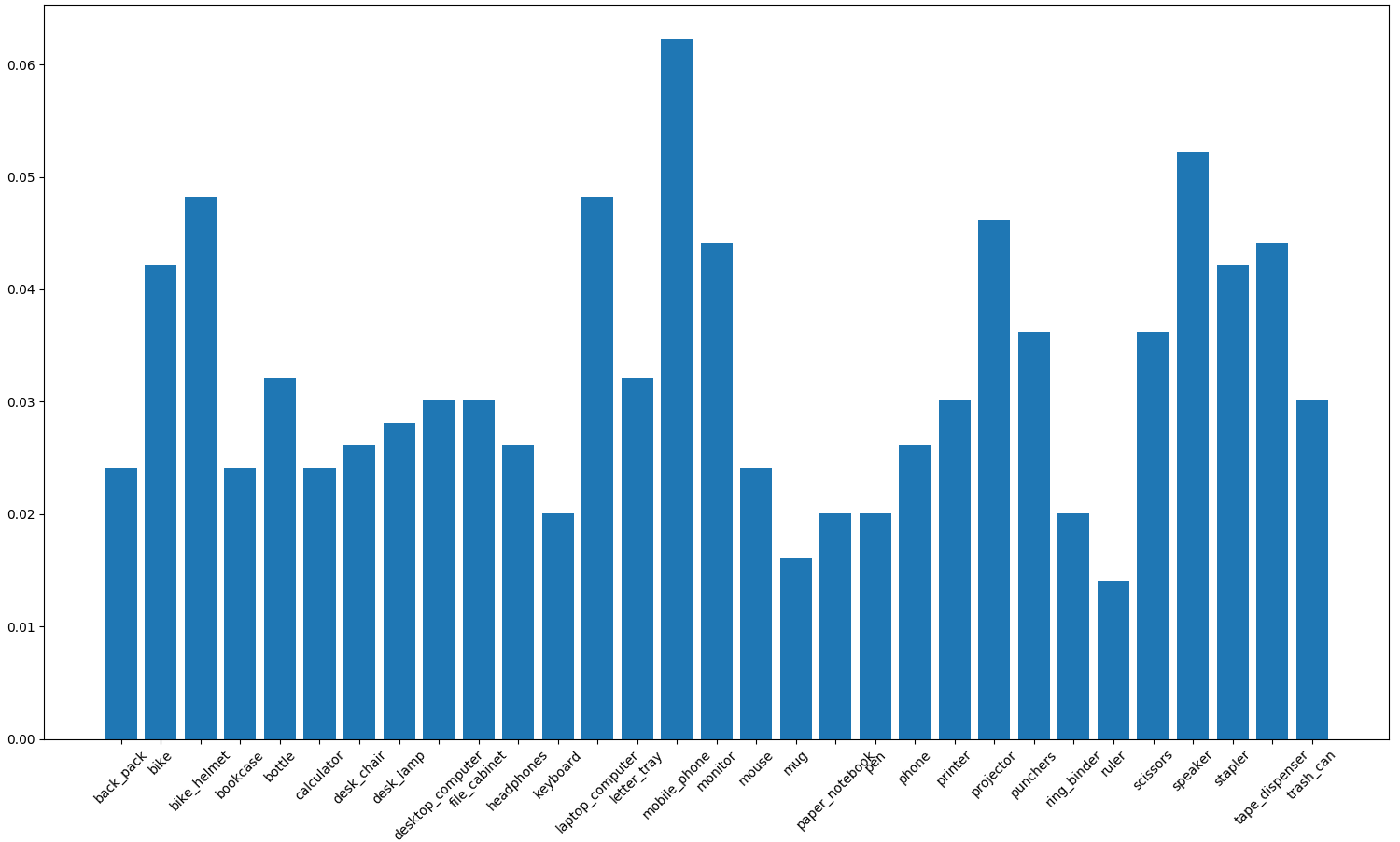}
    \caption{DSLR}
  \end{subfigure}
  \vskip\baselineskip
  \begin{subfigure}[b]{0.35\textwidth}   
    \centering 
    \includegraphics[width=\textwidth]{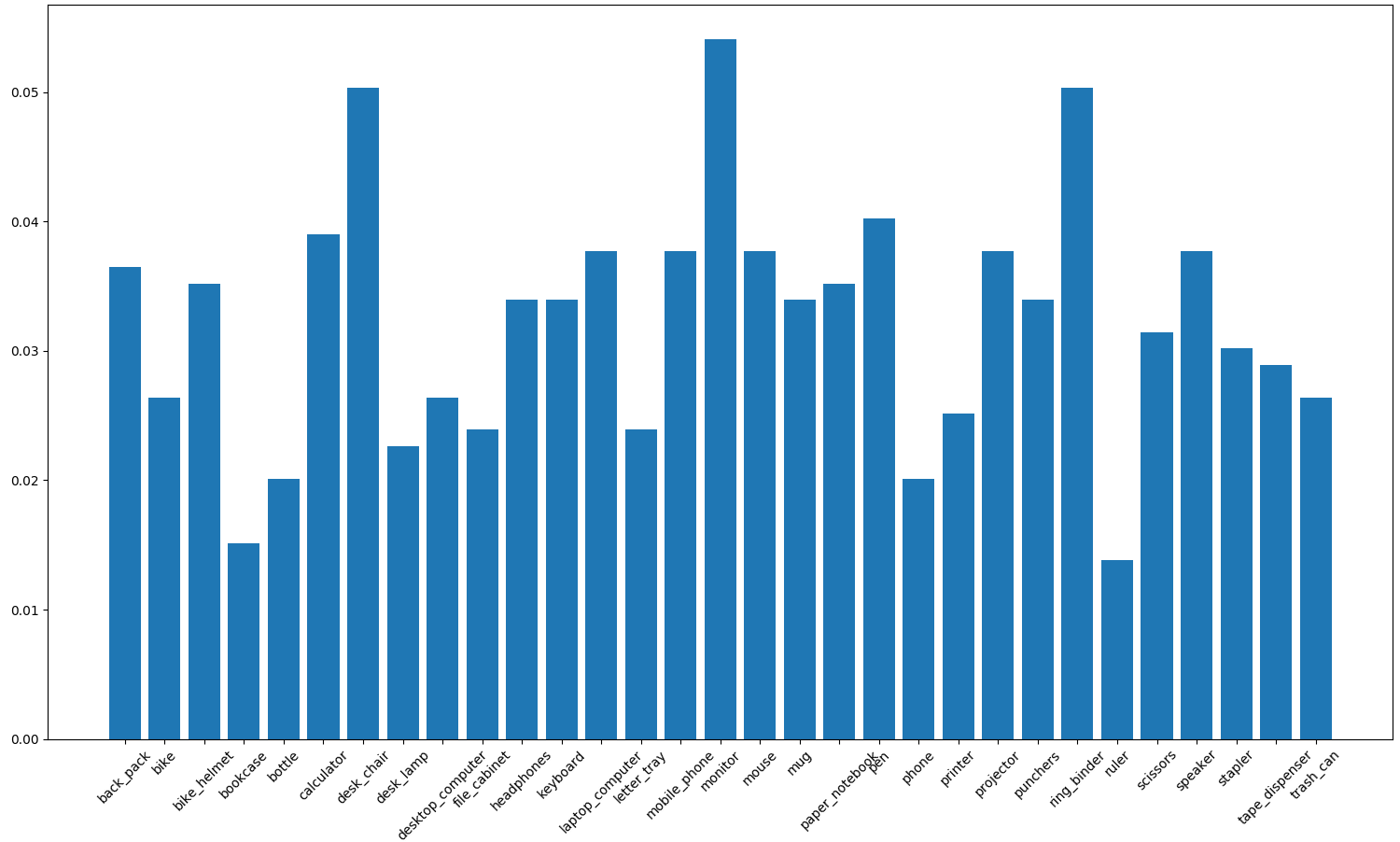}
    \caption{Webcam}
  \end{subfigure}
  \caption{Label distributions in Office-31.} 
  \label{fig:office_dist}
\end{figure}

\begin{figure}
  \centering
  \begin{subfigure}[b]{0.24\textwidth}
    \centering
    \includegraphics[width=\textwidth]{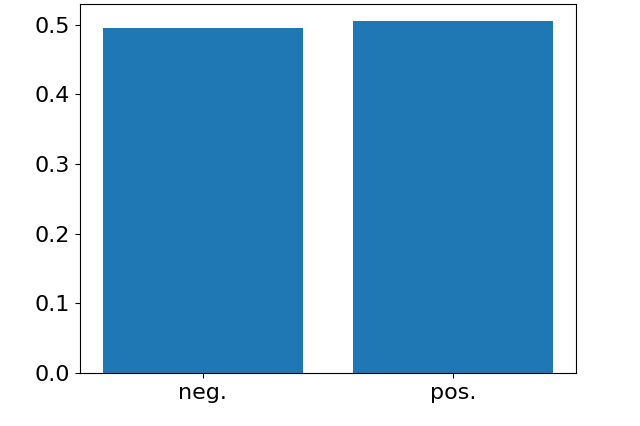}
    \caption{Books}
  \end{subfigure}
  \hfill
  \begin{subfigure}[b]{0.24\textwidth}  
    \centering 
    \includegraphics[width=\textwidth]{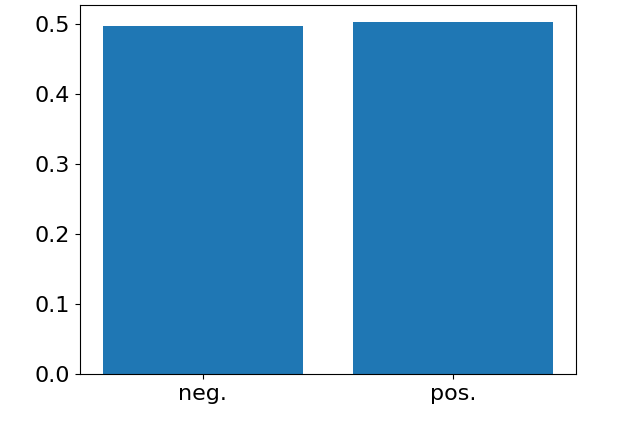}
    \caption{DVD}
  \end{subfigure}
  \hfill
  \begin{subfigure}[b]{0.24\textwidth}   
    \centering 
    \includegraphics[width=\textwidth]{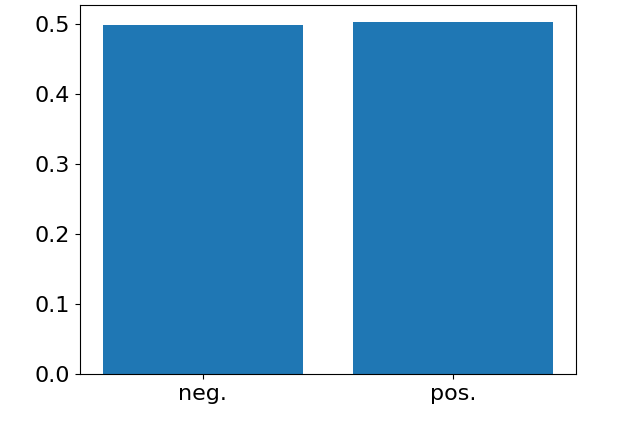}
    \caption{Electronics}
  \end{subfigure}
  \hfill
  \begin{subfigure}[b]{0.24\textwidth}   
    \centering 
    \includegraphics[width=\textwidth]{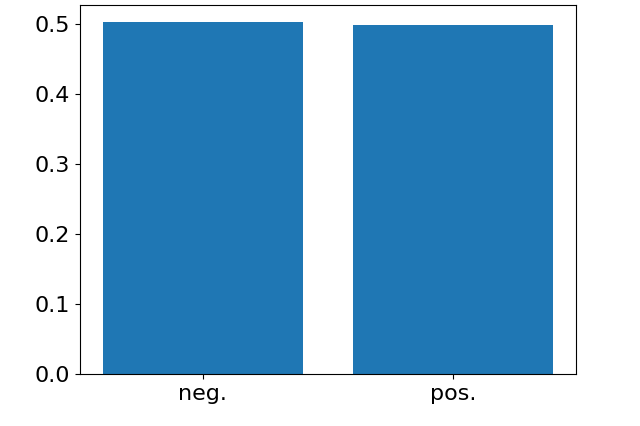}
    \caption{Kitchen}
  \end{subfigure}
  \caption{Label distributions in Amazon Reviews.} 
  \label{fig:amazon_dist}
\end{figure}

As remarked throughout this work and formally discussed and proven in \cite{Zhao2019}, having different marginal label distributions across source and target domains poses difficulties to the DA task and, under this scenario, domain-invariant representations can increase the error on the target domain. Since we claim that the adopted consistency regularization can help mitigate this issue, it is relevant to observe the distributions of labels across domains in the datasets used in this work, which are shown in Figures
~\ref{fig:digits_dist},~\ref{fig:office_dist}~and~\ref{fig:amazon_dist}. We also provide the pairwise Jensen-Shannon distances for the label distributions of all domains in Tables~\ref{tab:digits_js},~\ref{tab:office_js}~and~\ref{tab:amazon_js}. The Jensen-Shannon distance is convenient since it is a metric (so it is symmetric,  non-negative and zero if and only if the two distributions coincide) and upper bounded by $\sqrt{\log(2)}\,(\approx 0.8326)$. Therefore, it provides a comprehensive measure to evaluate the dissimilarity between two distributions.

In the digits datasets~(\Figref{fig:digits_dist}, \Tableref{tab:digits_js}), MNIST and MNIST-M have very similar label distributions and SynthDigits has an almost uniform label distribution. SVHN has the most skewed distribution, which is radically different from the remaining. Thus, this is the most challenging domain from this perspective. In Office-31~(\Figref{fig:office_dist}, \Tableref{tab:office_js}), label distributions across domains differ significantly. In the Amazon Reviews dataset~(\Figref{fig:amazon_dist}, \Tableref{tab:amazon_js}), the label distribution is almost uniform for all domains.

\begin{table}
\centering
\caption{Jensen-Shannon distances between the label distributions of each pair of digits domains. On each column, the largest distance is in bold and the smallest is underlined.}
\small
\begin{tabular}{l|cccc}
            & MNIST                & MNIST-M              & SVHN & SynthDigits \\ \hline
MNIST       & $0$                  & $\ut{2.73 \cdot 10^{-4}}$ & $\ut{1.17 \cdot 10^{-1}}$     & $\ut{1.83 \cdot 10^{-2}}$            \\
MNIST-M     & $\ut{2.73 \cdot 10^{-4}}$ & $0$                  & $\ut{1.17 \cdot 10^{-1}}$     & $1.84 \cdot 10^{-2}$            \\
SVHN        & $\bs{1.17 \cdot 10^{-1}}$ & $\bs{1.17 \cdot 10^{-1}}$ & $0$                      & $\bs{1.26 \cdot 10^{-1}}$            \\
SynthDigits & $1.83 \cdot 10^{-2}$ & $1.84 \cdot 10^{-2}$ & $\bs{1.26 \cdot 10^{-1}}$     & $0$ \\ \hline
Average & $4.51 \cdot 10^{-2}$ &  $4.52 \cdot 10^{-2}$ & $1.20 \cdot 10^{-1}$ & $5.42 \cdot 10^{-2}$
\end{tabular}
\label{tab:digits_js}
\end{table}

\begin{table}
\centering
\caption{Jensen-Shannon distances between the label distributions of each pair of domains in Office-31. On each column, the largest distance is in bold.}
\small
\begin{tabular}{l|ccc}
       & Amazon                    & DSLR                      & Webcam                    \\ \hline
Amazon & 0                         & $1.33 \cdot 10^{-1}$      & $9.76 \cdot 10^{-2}$      \\
DSLR   & $\bs{1.33 \cdot 10^{-1}}$ & 0                         & $\bs{1.45 \cdot 10^{-1}}$ \\
Webcam & $9.76 \cdot 10^{-2}$      & $\bs{1.45 \cdot 10^{-1}}$ & 0 \\ \hline
Average & $1.15 \cdot 10^{-1}$ & $1.39 \cdot 10^{-1}$ & $1.21 \cdot 10^{-1}$
\end{tabular}
\label{tab:office_js}
\end{table}

\begin{table}
\centering
\caption{Jensen-Shannon distances between the label distributions of each pair of domains in Amazon Reviews. On each column, the largest distance is in bold and the smallest is underlined.}
\small
\begin{tabular}{l|cccc}
            & Books                & DVD                  & Electronics          & Kitchen              \\ \hline
Books       & 0                    & $1.67 \cdot 10^{-3}$ & $1.93 \cdot 10^{-3}$ & $\bs{5.09 \cdot 10^{-3}}$ \\
DVD         & $\ut{1.67 \cdot 10^{-3}}$ & 0                    & $\ut{2.53 \cdot 10^{-4}}$ & $3.42 \cdot 10^{-3}$ \\
Electronics & $1.93 \cdot 10^{-3}$ & $\ut{2.53 \cdot 10^{-4}}$ & 0                    & $\ut{3.17 \cdot 10^{-3}}$ \\
Kitchen     & $\bs{5.09 \cdot 10^{-3}}$ & $\bs{3.42 \cdot 10^{-3}}$ & $\bs{3.17 \cdot 10^{-3}}$ & 0 \\ \hline
Average & $2.90 \cdot 10^{-3}$ & $1.78 \cdot 10^{-3}$ & $1.78 \cdot 10^{-3}$ & $3.89 \cdot 10^{-3}$
\end{tabular}
\label{tab:amazon_js}
\end{table}

\subsection{Effect of over-training}
\label{sec:overtrain}
When the label distributions differ across domains, training the feature extractor and domain discriminator for a large number of iterations tends to lead to an increased target error. This is a direct effect of the curse of domain-invariant representations and it has been verified experimentally in \cite{Zhao2019}.

\begin{figure}
\centering
\begin{subfigure}[b]{0.32\textwidth}
\begin{tikzpicture}[scale=0.58, every node/.style={scale=0.58}]
\begin{axis}[
    enlargelimits=false,
    legend style={at={(0.45,0.1)},anchor=south west,font=\large},
    xlabel=\huge epoch, xmin=0, xmax=60, xtick={0,10,20,30,40,50,60},
    ylabel=\huge accuracy, ymin=0.55, ymax=0.75, ytick={0.55,0.6,0.65,0.7,0.75},
    grid=both, grid style={line width=.1pt, draw=gray!10},
]   
    \addplot[color=myblue,smooth,thick]table[x=Epoch, y=MDAN, col sep=comma]{data/amazon_long.csv};
    \addlegendentry{MDAN~\cite{Zhao2018} ($\pgfmathprintnumber{\slopeMDAN}$)}

    \addplot[color=myorange,smooth,thick]table[x=Epoch, y=MixMDAN, col sep=comma]{data/amazon_long.csv};
    \addlegendentry{MODA ($\pgfmathprintnumber{\slopeMODA}$)}
    
    \addplot[color=mygreen,smooth,thick]table[x=Epoch, y=Ours, col sep=comma]{data/amazon_long.csv};
    \addlegendentry{MODA-FM ($\pgfmathprintnumber{\slopeMODAFM}$)}
    
    \addplot[color=gray,dashed] table [col sep = comma,y={create col/linear regression={y=1}},skip first n=2]{data/amazon_long.csv};
    \xdef\slopeMDAN{\pgfplotstableregressiona}
    \addplot[color=gray,dashed] table [col sep = comma,y={create col/linear regression={y=2}},skip first n=2]{data/amazon_long.csv};
    \xdef\slopeMODA{\pgfplotstableregressiona}
    \addplot[color=gray,dashed] table [col sep = comma,y={create col/linear regression={y=3}},skip first n=2]{data/amazon_long.csv};
    \xdef\slopeMODAFM{\pgfplotstableregressiona}
\end{axis}
\end{tikzpicture}
\caption{Amazon}
\end{subfigure}
\hfill%
\begin{subfigure}[b]{0.32\textwidth}
\centering
\begin{tikzpicture}[scale=0.58, every node/.style={scale=0.58}]
\begin{axis}[
    enlargelimits=false,
    legend style={at={(0.45,0.1)},anchor=south west,font=\large},
    xlabel=\huge epoch, xmin=0, xmax=60, xtick={0,10,20,30,40,50,60},
    ylabel=\huge accuracy, ymin=0.8, ymax=1, ytick={0.8,0.85,0.9,0.95,1},
    grid=both, grid style={line width=.1pt, draw=gray!10},
]   
    \addplot[color=myblue,smooth,thick]table[x=Epoch, y=MDAN, col sep=comma]{data/dslr_long.csv};
    \addlegendentry{MDAN~\cite{Zhao2018} ($\pgfmathprintnumber{\slopeMDAN}$)}

    \addplot[color=myorange,smooth,thick]table[x=Epoch, y=MixMDAN, col sep=comma]{data/dslr_long.csv};
    \addlegendentry{MODA ($\pgfmathprintnumber{\slopeMODA}$)}
    
    \addplot[color=mygreen,smooth,thick]table[x=Epoch, y=Ours, col sep=comma]{data/dslr_long.csv};
    \addlegendentry{MODA-FM ($\pgfmathprintnumber{\slopeMODAFM}$)}
    
    \addplot[color=gray,dashed] table [col sep = comma,y={create col/linear regression={y=1}},skip first n=2]{data/dslr_long.csv};
    \xdef\slopeMDAN{\pgfplotstableregressiona}
    \addplot[color=gray,dashed] table [col sep = comma,y={create col/linear regression={y=2}},skip first n=2]{data/dslr_long.csv};
    \xdef\slopeMODA{\pgfplotstableregressiona}
    \addplot[color=gray,dashed] table [col sep = comma,y={create col/linear regression={y=3}},skip first n=2]{data/dslr_long.csv};
    \xdef\slopeMODAFM{\pgfplotstableregressiona}
\end{axis}
\end{tikzpicture}
\caption{DSLR}
\end{subfigure}
\hfill%
\begin{subfigure}[b]{0.32\textwidth}
\centering
\begin{tikzpicture}[scale=0.58, every node/.style={scale=0.58}]
\begin{axis}[
    enlargelimits=false,
    legend style={at={(0.45,0.1)},anchor=south west,font=\large},
    xlabel=\huge epoch, xmin=0, xmax=60, xtick={0,10,20,30,40,50,60},
    ylabel=\huge accuracy, ymin=0.8, ymax=1, ytick={0.8,0.85,0.9,0.95,1},
    grid=both, grid style={line width=.1pt, draw=gray!10},
]   
    \addplot[color=myblue,smooth,thick]table[x=Epoch, y=MDAN, col sep=comma]{data/webcam_long.csv};
    \addlegendentry{MDAN~\cite{Zhao2018} ($\pgfmathprintnumber{\slopeMDAN}$)}

    \addplot[color=myorange,smooth,thick]table[x=Epoch, y=MixMDAN, col sep=comma]{data/webcam_long.csv};
    \addlegendentry{MODA ($\pgfmathprintnumber{\slopeMODA}$)}
    
    \addplot[color=mygreen,smooth,thick]table[x=Epoch, y=Ours, col sep=comma]{data/webcam_long.csv};
    \addlegendentry{MODA-FM ($\pgfmathprintnumber{\slopeMODAFM}$)}
    
    \addplot[color=gray,dashed] table [col sep = comma,y={create col/linear regression={y=1}},skip first n=2]{data/webcam_long.csv};
    \xdef\slopeMDAN{\pgfplotstableregressiona}
    \addplot[color=gray,dashed] table [col sep = comma,y={create col/linear regression={y=2}},skip first n=2]{data/webcam_long.csv};
    \xdef\slopeMODA{\pgfplotstableregressiona}
    \addplot[color=gray,dashed] table [col sep = comma,y={create col/linear regression={y=3}},skip first n=2]{data/webcam_long.csv};
    \xdef\slopeMODAFM{\pgfplotstableregressiona}
\end{axis}
\end{tikzpicture}
\caption{Webcam}
\end{subfigure}
\caption{Test accuracy along 60 training epochs for our model and two baselines in Office-31. The tendency line for each curve is also shown (dashed lines) and the respective slope is indicated in brackets in each plot legend. The domain indicated below each plot is the target.}
\label{fig:overtrain}
\end{figure}
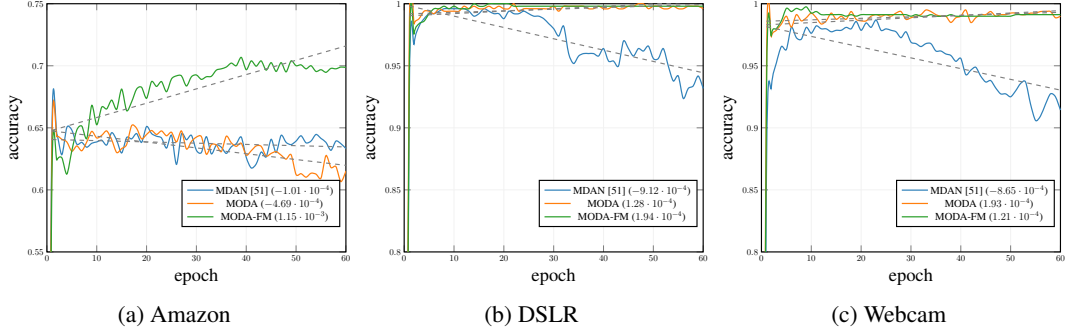

In this experiment, we want to evaluate if the increased robustness of the feature extractor provided by the adopted consistency regularization helps to mitigate this issue. For this purpose, we use Office-31, as this is the dataset where the label distributions are most different across domains (see~\secref{sec:label_dist}). We train our model and two baselines (MDAN~\cite{Zhao2018} and MODA) for 60 epochs and we observe the evolution of the model accuracy on the target data. The plots are shown in \Figref{fig:overtrain}. As we see there, in MODA-FM the accuracy keeps stable after reaching the maximum in all domains. Contrarily, in MODA and especially in MDAN, the accuracy tends to decay after reaching the maximum. These observations strongly suggest that MODA-FM succeeds on mitigating the curse of domain-invariant representations. In MODA the problem is less pronounced than in MDAN probably because the latter will continue optimizing itself until it can produce domain-invariant representations for all source domains, whereas the former will simply ignore the hardest source domains by assigning them a low weight (possibly even zero).

\subsection{Hyperparameter sensitivity analysis}
\label{sec:hyperparam}
Our model comprises three main hyperparameters: $\mu_d$, $\mu_s$ and $\mu_c$. The hyperparameter $\mu_d$ controls the relative weight of the domain discriminator loss and is present in every work on adversarial DA, thus its effect has already been studied extensively. For this reason, we focus on the effect of $\mu_s$ and $\mu_c$. We use the digits datasets and evaluate the accuracy on each target domain while varying either $\mu_s$ or $\mu_c$ and keeping the other one constant at the optimal value.

\begin{figure}
\centering
\begin{subfigure}[b]{0.45\textwidth}
\centering
\begin{tikzpicture}[scale=0.8, every node/.style={scale=0.58}]
\begin{semilogxaxis}[xmin=5e-5, xmax=50, ymin=0, ymax=1.05, domain=1e-5:20, ylabel=\large accuracy,
    scaled x ticks=real:1e-3,
    xtick scale label code/.code={},
    log x ticks with fixed point/.style={
      xticklabel={
        \pgfkeys{/pgf/fpu=true}
        \pgfmathparse{exp(\tick)}%
        \pgfmathprintnumber[fixed relative, precision=3]{\pgfmathresult}
        \pgfkeys{/pgf/fpu=false}
      }
    },
    legend style={at={(0.65,0.1)},anchor=south west,font=\normalsize},
    grid=both, grid style={line width=.1pt, draw=gray!10}]
    
    \addplot[color=myblue,only marks,sharp plot,mark=*]table[x index={0}, y index={1}, col sep=comma, skip first n=2]{data/mu_s.csv};
    \addlegendentry{MNIST}
    
    \addplot[color=myorange,only marks,sharp plot,mark=*]table[x index={0}, y index={2}, col sep=comma, skip first n=2]{data/mu_s.csv};
    \addlegendentry{MNIST-M}
    
    \addplot[color=mygreen,only marks,sharp plot,mark=*]table[x index={0}, y index={3}, col sep=comma, skip first n=2]{data/mu_s.csv};
    \addlegendentry{SVHN}
\end{semilogxaxis}
\end{tikzpicture}
\caption{$\mu_s$}
\label{fig:hyperparam_mu_s}
\end{subfigure}
\hfill%
\begin{subfigure}[b]{0.45\textwidth}
\centering
\begin{tikzpicture}[scale=0.8, every node/.style={scale=0.58}]
\begin{semilogxaxis}[xmin=5e-5, xmax=50, ymin=0, ymax=1.05, domain=1e-5:20, ylabel=\large accuracy,
    scaled x ticks=real:1e-3,
    xtick scale label code/.code={},
    log x ticks with fixed point/.style={
      xticklabel={
        \pgfkeys{/pgf/fpu=true}
        \pgfmathparse{exp(\tick)}%
        \pgfmathprintnumber[fixed relative, precision=3]{\pgfmathresult}
        \pgfkeys{/pgf/fpu=false}
      }
    },
    legend style={at={(0.65,0.1)},anchor=south west,font=\normalsize},
    grid=both, grid style={line width=.1pt, draw=gray!10}]
    
    \addplot[color=myblue,only marks,sharp plot,mark=*]table[x index={0}, y index={1}, col sep=comma, skip first n=2]{data/mu_c.csv};
    \addlegendentry{MNIST}
    
    \addplot[color=myorange,only marks,sharp plot,mark=*]table[x index={0}, y index={2}, col sep=comma, skip first n=2]{data/mu_c.csv};
    \addlegendentry{MNIST-M}
    
    \addplot[color=mygreen,only marks,sharp plot,mark=*]table[x index={0}, y index={3}, col sep=comma, skip first n=2]{data/mu_c.csv};
    \addlegendentry{SVHN}
\end{semilogxaxis}
\end{tikzpicture}
\caption{$\mu_c$}
\label{fig:hyperparam_mu_c}
\end{subfigure}
\caption{Test accuracy as a function of hyperparameters $\mu_s$ (a) and $\mu_c$ (b) using the digits datasets. The domain corresponding to each line is the target.}
\label{fig:hyperparam}
\end{figure}
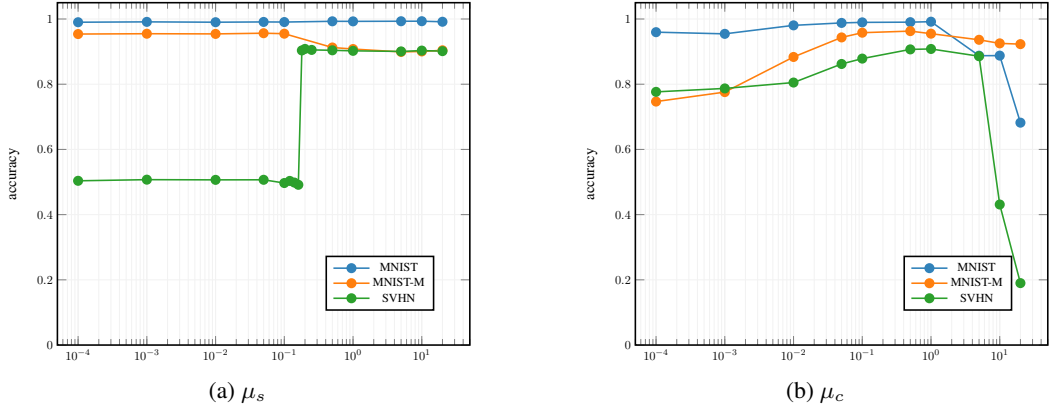

A sufficiently large value of $\mu_s$ forces $\alpha$ to converge to a vector with all components equal to $1/M$, weighting all source domains equally. Setting $\mu_s$ close to zero corresponds to the most optimistic scenario: the sparsity penalization is dropped and so, after sufficiently many training iterations, $\alpha$ would converge to a one-hot vector, choosing the source domain with the minimum difference of classification and discrimination losses ($\gL_{class} - \mu_d\gL_{disc}$). \Figref{fig:hyperparam_mu_s} shows that, for MNIST, similar results are obtained with any of those strategies. The fact that digits classification in MNIST is significantly easier than in any of the remaining datasets likely explains this behavior. Domain adaptation for MNIST-M is slightly favored by smaller values of $\mu_s$, meaning that a more optimistic approach works better here. For SVHN, the opposite holds and the results are very poor when low sparsity regularization is employed. There is even a strong discontinuity in the accuracy plot for this domain around $\mu_s \approx 0.2$. This discontinuity divides situations where $\alpha$ collapses into the easiest source domain ($\mu_s < 0.2$) and those where it does not collapse ($\mu_s > 0.2$). We elaborate more on this in \secref{sec:alpha_evol}. In all cases, we observe that a $\mu_s \in [0.2, 0.6]$ provides near-optimal results.

The effect of varying $\mu_c$ is plotted in \Figref{fig:hyperparam_mu_c}. Values of $\mu_c$ close to zero drop the consistency regularization and, therefore, we obtain results close to those of MODA. Significant performance gains are obtained for $\mu_c > 0.01$ and the optimal is reached for $\mu_c \in [0.4, 1.0]$ for all domains.

\subsection{Evolution of the source weights}
\label{sec:alpha_evol}
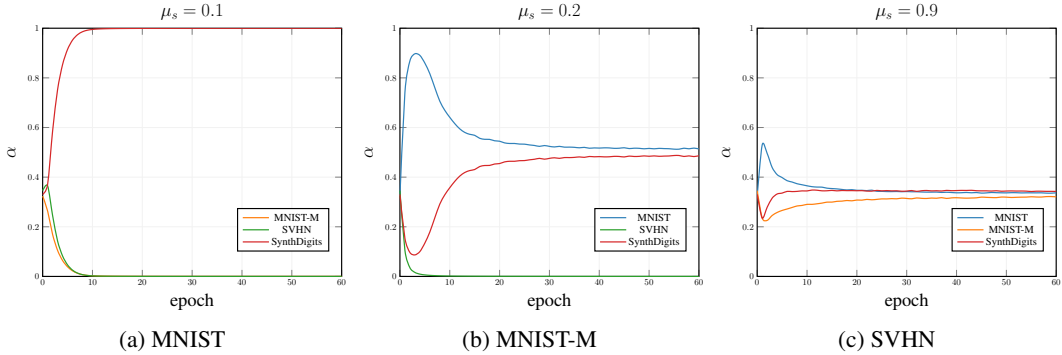
\begin{figure}
\centering
\begin{subfigure}[b]{0.32\textwidth}
\begin{tikzpicture}[scale=0.58, every node/.style={scale=0.58}]
\begin{axis}[
    enlargelimits=false,
    legend style={at={(0.65,0.1)},anchor=south west,font=\large},
    xlabel=\huge epoch, xmin=0, xmax=60, xtick={0,10,20,30,40,50,60},
    ylabel=\huge $\alpha$, ymin=0., ymax=1., 
    grid=both, grid style={line width=.1pt, draw=gray!10},
    title=\text{\huge $\mu_s=0.1$}
]   
    \addplot[color=myorange,smooth,thick]table[x=Epoch, y=MNIST_M, col sep=comma]{data/mnist_coef.csv};
    \addlegendentry{MNIST-M}

    \addplot[color=mygreen,smooth,thick]table[x=Epoch, y=SVHN, col sep=comma]{data/mnist_coef.csv};
    \addlegendentry{SVHN}
    
    \addplot[color=myred,smooth,thick]table[x=Epoch, y=SynthDigits, col sep=comma]{data/mnist_coef.csv};
    \addlegendentry{SynthDigits}
\end{axis}
\end{tikzpicture}
\caption{MNIST}
\end{subfigure}
\hfill%
\begin{subfigure}[b]{0.32\textwidth}
\centering
\begin{tikzpicture}[scale=0.58, every node/.style={scale=0.58}]
\begin{axis}[
    enlargelimits=false,
    legend style={at={(0.65,0.1)},anchor=south west,font=\large},
    xlabel=\huge epoch, xmin=0, xmax=60, xtick={0,10,20,30,40,50,60},
    ylabel=\huge $\alpha$, ymin=0., ymax=1.,
    grid=both, grid style={line width=.1pt, draw=gray!10},
    title=\text{\huge$\mu_s=0.2$}
]   
    \addplot[color=myblue,smooth,thick]table[x=Epoch, y=MNIST, col sep=comma]{data/mnist_m_coef.csv};
    \addlegendentry{MNIST}

    \addplot[color=mygreen,smooth,thick]table[x=Epoch, y=SVHN, col sep=comma]{data/mnist_m_coef.csv};
    \addlegendentry{SVHN}
    
    \addplot[color=myred,smooth,thick]table[x=Epoch, y=SynthDigits, col sep=comma]{data/mnist_m_coef.csv};
    \addlegendentry{SynthDigits}
\end{axis}
\end{tikzpicture}
\caption{MNIST-M}
\end{subfigure}
\hfill%
\begin{subfigure}[b]{0.32\textwidth}
\centering
\begin{tikzpicture}[scale=0.58, every node/.style={scale=0.58}]
\begin{axis}[
    enlargelimits=false,
    legend style={at={(0.65,0.1)},anchor=south west,font=\large},
    xlabel=\huge epoch, xmin=0, xmax=60, xtick={0,10,20,30,40,50,60},
    ylabel=\huge $\alpha$, ymin=0., ymax=1,
    grid=both, grid style={line width=.1pt, draw=gray!10},
    title=\text{\huge$\mu_s=0.9$}
]   
    \addplot[color=myblue,smooth,thick]table[x=Epoch, y=MNIST, col sep=comma]{data/svhn_coef.csv};
    \addlegendentry{MNIST}

    \addplot[color=myorange,smooth,thick]table[x=Epoch, y=MNIST_M, col sep=comma]{data/svhn_coef.csv};
    \addlegendentry{MNIST-M}
    
    \addplot[color=myred,smooth,thick]table[x=Epoch, y=SynthDigits, col sep=comma]{data/svhn_coef.csv};
    \addlegendentry{SynthDigits}
\end{axis}
\end{tikzpicture}
\caption{SVHN}
\end{subfigure}
\caption{Source weights $\alpha$ for each domain along 60 training epochs in the digits datasets. The target domain and the value of $\mu_s$ that was used are indicated below and above each plot, respectively.}
\label{fig:alpha_train}
\end{figure}

Here, we study the evolution of the source weights $\alpha$ along training. The behavior is determined by the evolution of the classification and discrimination losses of each source domain and also by the choice of the hyperparameter $\mu_s$. The effect of varying $\mu_s$ on the model performance was analyzed in \secref{sec:hyperparam}. Now, we are interested in observing the dynamics of $\alpha$ along the training epochs, for different source domains, using the corresponding optimal $\mu_s$. We use the digits datasets for this purpose and we present the results in \Figref{fig:alpha_train}.

For MNIST, where the lowest $\mu_s$ is used, we observe that the mixture coefficients rapidly collapse into the easiest source domain (SynthDigits). Nonetheless, in \secref{sec:hyperparam} we have observed that the target accuracy for MNIST was almost insensitive to the choice of $\mu_s$, meaning that using only the data from the easiest source domain or weighting all source domains equally would produce identical results. When we take MNIST-M and SVHN as target domains, with a slightly increased $\mu_s$, a different behavior occurs. Early in the training process, the weight for the easiest source domain (MNIST) increases rapidly, following the fast decrease in the corresponding loss. Later, as the loss for the easiest source domain plateaus and the remaining keep decreasing, the  weights for the remaining active source domains start to increase. This behavior explains the discontinuity we have observed in the plot of target accuracy vs. $\mu_s$ for SVHN (\Figref{fig:hyperparam_mu_s}): if $\mu_s$ is sufficiently small to allow $\alpha$ to rapidly collapse into MNIST, the data of the remaining source domains is simply discarded for the remaining of the training process and the corresponding weights will never increase.

\subsection{Network architectures}
\label{sec:arch}
The following notation is used to designate network layers: \texttt{Conv(n, k)} -- 2-D convolutional layer with \texttt{n} output channels, square kernels with size \texttt{k}, and unit stride, followed by a rectified linear activation; \texttt{MaxPool(k)} -- 2-D max-pooling over non-overlapping regions of $\texttt{k} \times \texttt{k}$ pixels; \texttt{AdaptAvgPool(k)} -- 2-D adaptive average pooling where the output size is $\texttt{k} \times \texttt{k}$ pixels. \texttt{FC(n)} -- fully-connected layer with $\texttt{n}$ output neurons, followed by a rectified linear activation except for output layers; \texttt{Dropout(p)} -- dropout layer where $\texttt{p}$ is the probability of zeroing an element. A gradient reversal layer~\cite{Ganin2015} is present at the input of the domain discriminator in all models. All classifiers and domain discriminators are followed by a softmax layer that maps the output to the corresponding class probabilities.

\paragraph{Digits classification} Input images have shape $3 \times 32 \times 32$. Feature extractor: \texttt{Conv(64, 3)}~$\rightarrow$ \texttt{MaxPool(2)}~$\rightarrow$ \texttt{Conv(128, 3)}~$\rightarrow$ \texttt{MaxPool(2)}~$\rightarrow$ \texttt{Conv(256, 3)}. Task classifier: \texttt{MaxPool(2)}~$\rightarrow$ \texttt{Conv(256, 3)}~$\rightarrow$ \texttt{FC(2048)}~$\rightarrow$ \texttt{FC(1024)}~$\rightarrow$ \texttt{FC(10)}. Domain discriminator: \texttt{MaxPool(2)}~$\rightarrow$ \texttt{FC(2048)}~$\rightarrow$ \texttt{FC(2048)}~$\rightarrow$ \texttt{FC(1024)}~$\rightarrow$ \texttt{FC(2)}.

\paragraph{Object classification (Office-31 dataset)} Input images have shape $3 \times 256 \times 256$. Feature extractor: ResNet-50 up to stage~3. Task classifier / domain discriminator: ResNet-50 stage~4~$\rightarrow$ \texttt{AdaptAvgPool(1)}~$\rightarrow$ \texttt{FC(\#classes)}, where $\texttt{\#classes} = 31$ for the task classifier and $\texttt{\#classes} = 2$ for the domain discriminator.

\paragraph{Sentiment analysis (Amazon Reviews dataset)} Input features have dimension $5000$. Feature extractor: \texttt{Dropout(p)}~$\rightarrow$ \texttt{FC(1000)}~$\rightarrow$ \texttt{Dropout(p)}~$\rightarrow$ \texttt{FC(500)}~$\rightarrow$ \texttt{Dropout(p)}~$\rightarrow$ \texttt{FC(100)}~$\rightarrow$ \texttt{Dropout(p)}, where $\texttt{p} = 0$ except for producing the augmented samples used in the consistency regularization. Task classifier / domain discriminator: \texttt{FC(2)}.

\subsection{Choice of hyperparameters}
\label{sec:cross_val}
\Tableref{tab:hyperparameters} presents the values assigned to each hyperparameter. For some hyperparameters, a set or range of values is presented instead of a single value, meaning that the value for that hyperparameter was selected via cross-validation on the given set/range. This cross-validation consisted on 20 iterations of random search for each experiment, where no data from the target domain was used and each source domain was taken as target in turn. The hyperparameter combination that yielded the best average result was chosen.

\begin{table}
\centering
\caption{Values and search ranges for all hyperparameters in all experiments.}
\label{tab:hyperparameters}
\small
\begin{tabular}{r|ccc}
\multicolumn{1}{l|}{Hyperparameter} & Digits                & Office-31             & Amazon Reviews          \\ \hline
$\mu_d$~(8)                            & $[10^{-4}, 10^{0}]$   & $[10^{-4}, 10^{0}]$   & $[10^{-4}, 10^{0}]$     \\ \hdashline[0.5pt/5pt]
$\mu_s$~(8)                             & $[10^{-5}, 10^{0}]$   & $10^{-2}$             & $[10^{-6}, 10^{-1}]$    \\ \hdashline[0.5pt/5pt]
$\mu_c$~(16)                             & $[10^{-2}, 10^{0}]$   & $[10^{-2}, 10^{0}]$   & $[10^{-2}, 10^{0}]$     \\ \hdashline[0.5pt/5pt]
$\tau$~(15)                              & $0.9$                 & $0.9$                 & $0.9$                   \\ \hdashline[0.5pt/5pt]
optimizer                           & AdaDelta              & AdaDelta              & AdaDelta                \\ \hdashline[0.5pt/5pt]
no. epochs                          & $\{10, 20, \dots, 60\}$ & $\{10, 15, \dots, 60\}$ & $\{5, 10, \dots, 40\}$ \\ \hdashline[0.5pt/5pt]
learning rate                         & $0.1$             & $0.1$             & $1.0$                \\ \hdashline[0.5pt/5pt]
batch size                          & $8$                   & $8$                   & $20$                    \\ \hdashline[0.5pt/5pt]
RandAugment: $n$                    & $2$                   & $2$                   & n.a.                    \\ \hdashline[0.5pt/5pt]
RandAugment: $m_{\text{min}}$       & $3$                   & $3$                   & n.a.                    \\ \hdashline[0.5pt/5pt]
RandAugment: $m_{\text{max}}$       & $10$                  & $10$                  & n.a.                    \\ \hdashline[0.5pt/5pt]
dropout: $p_{\text{min}}$           & n.a.                  & n.a.                  & $0.2$                   \\ \hdashline[0.5pt/5pt]
dropout: $p_{\text{max}}$           & n.a.                  & n.a.                  & $0.8$                  
\end{tabular}
\end{table}

\subsection{Image transformations}
\label{sec:img_transf}
The list of image transformations used in RandAugment is provided in~\Tableref{tab:randaugment}. The cutout transformation is the first transformation to be applied to every image. This transformation occludes a randomly located square region with a length equal to $30\%$ of the image length. The remaining $n$ transformations are chosen at random from the provided list. All transformations were normalized such that a transformation of magnitude $m = 0$ corresponds to the identity (i.e., unchanged image) and a magnitude $m = 30$ (maximum admissible value) corresponds to the maximum distortion. We implemented the image transformations using the Python Imaging Library (PIL). 

\begin{savenotes}
\begin{table}[!ht]
\centering
\caption{RandAugment image transformations and respective ranges.}
\label{tab:randaugment}
\small
\begin{tabular}{r|c||rc}
\multicolumn{1}{c|}{Transformation} & Range         & \multicolumn{1}{c|}{Transformation} & \multicolumn{1}{c}{Range} \\ \hline
AutoContrast                        & $\{0, 1\}$    & \multicolumn{1}{r|}{Rotate}      & $[-30^{\circ}, 30^{\circ}]$                  \\ \hdashline[0.5pt/5pt]
Brightness                          & $[0.1, 1.9]$  & \multicolumn{1}{r|}{Sharpness}      & $[0.1, 1.9]$              \\ \hdashline[0.5pt/5pt]
Color                               & $[0.1, 1.9]$  & \multicolumn{1}{r|}{ShearX}         & $[0, 0.3]$                \\ \hdashline[0.5pt/5pt]
Contrast                            & $[0.1, 1.9]$  & \multicolumn{1}{r|}{ShearY}         & $[0, 0.3]$                \\ \hdashline[0.5pt/5pt]
Equalize                            & $\{0, 1\}$    & \multicolumn{1}{r|}{Solarize}       & $[0, 255]$                \\ \hdashline[0.5pt/5pt]
FlipX~\footnote{Excluded in the digits experiment since it is not label-preserving.}                  & $\{0, 1\}$    & \multicolumn{1}{r|}{TranslateX}     & $[-0.45, 0.45]$           \\ \hdashline[0.5pt/5pt]
Invert                              & $\{0, 1\}$    & \multicolumn{1}{r|}{TranslateY}     & $[-0.45, 0.45]$           \\ \hdashline[0.5pt/5pt]
Posterize                           & $[4, 8]$ &                                     &              
\end{tabular}
\end{table}%
\end{savenotes}

\subsection{Sample images}
\label{sec:sample_imgs}
\begin{figure}[h!]
  \includegraphics[width=\linewidth]{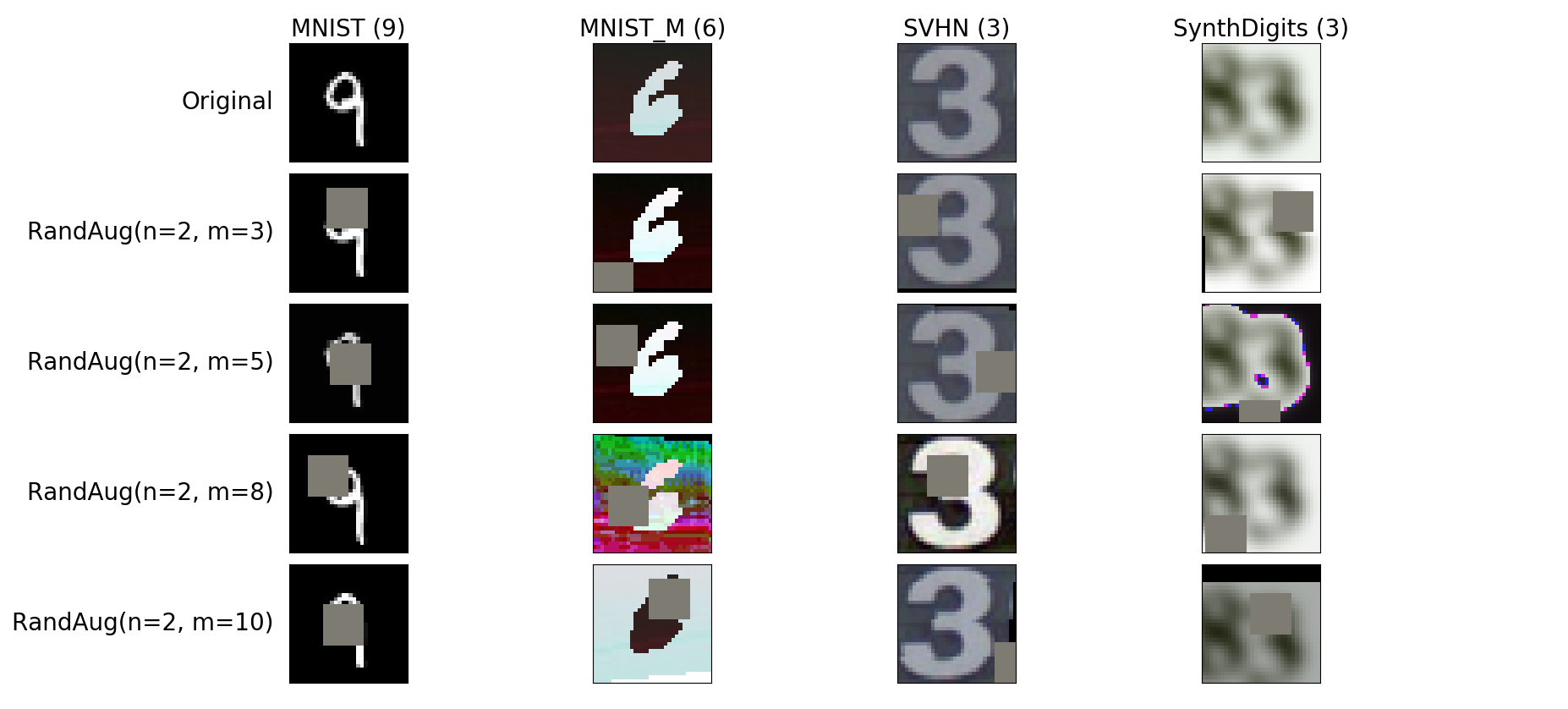}
  \caption{Sample original and transformed images from the digits datasets. The ground-truth label is in brackets.}
\end{figure}
\begin{figure}[h!]
  \includegraphics[width=\linewidth]{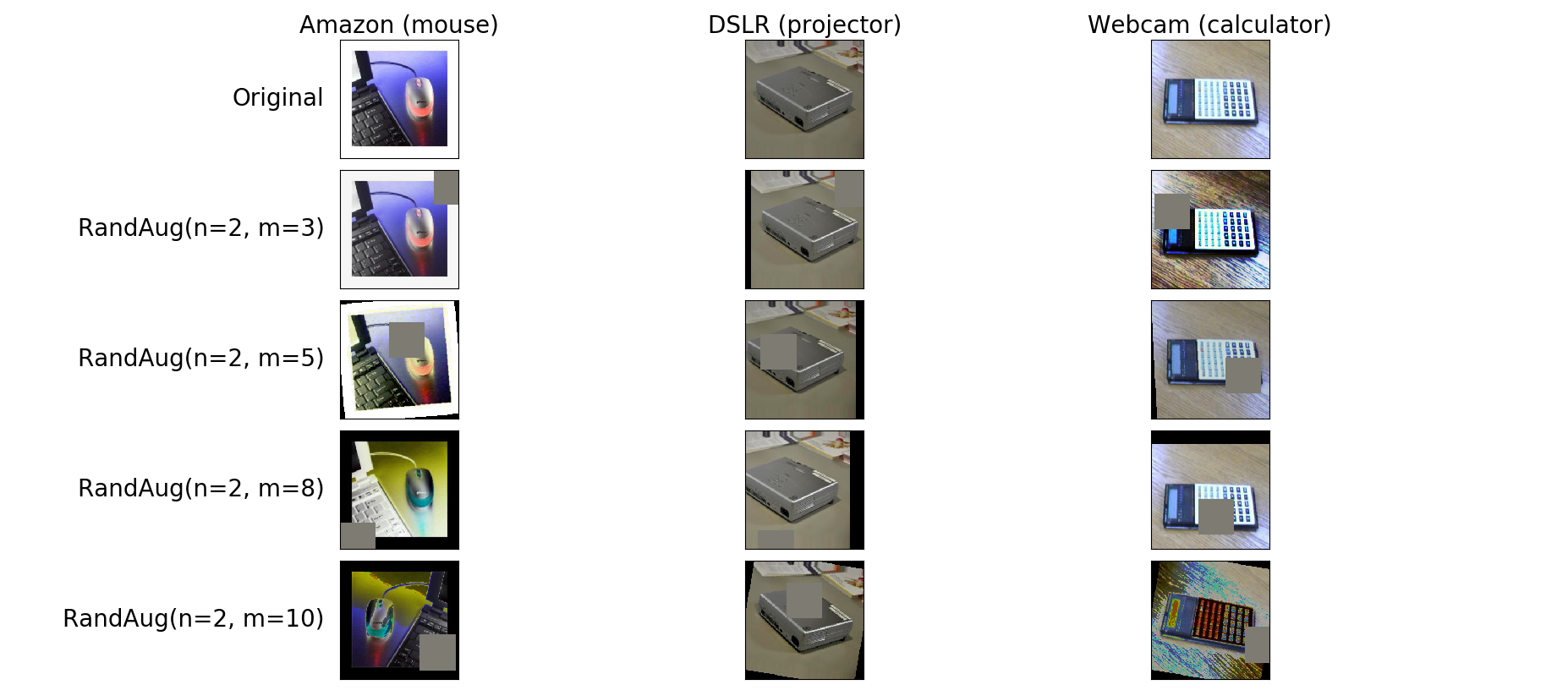}
  \caption{Sample original and transformed images from each domain in the Office-31 dataset. The ground-truth label is in brackets.}
\end{figure}

\end{document}